\documentclass{article} 
\usepackage{iclr2027_conference,times}

\newcommand{\llnote}[1]{\rlap{\textsubscript{\scriptsize~#1}}}
\newcommand{\std}[1]{\rlap{$\pm$#1}}
\newcommand{\sstd}[1]{\rlap{\scriptsize{$\pm$#1}}}

\usepackage{graphicx} 
\usepackage{booktabs}
\usepackage{multirow}
\usepackage{subcaption}
\usepackage{tabularx}
\usepackage[normalem]{ulem} 
\usepackage{amssymb}

\usepackage{amsmath,amsfonts,bm}

\def\eqref#1{equation~\ref{#1}}

\def\1{\bm{1}}

\DeclareMathAlphabet{\mathsfit}{\encodingdefault}{\sfdefault}{m}{sl}
\SetMathAlphabet{\mathsfit}{bold}{\encodingdefault}{\sfdefault}{bx}{n}

\usepackage{hyperref}
\usepackage{url}

\title{Teaching LLMs to See Graphs: Unifying Text and Structural Reasoning}

\author{%
  Dario Vajda\thanks{Work completed at the University of Ljubljana. The author is currently at the University of Cambridge.} \\
  University of Ljubljana \\
  University of Cambridge \\
  \texttt{vajdadario@gmail.com} \\
}

\iclrfinalcopy 
\begin{document}

\maketitle

\begin{abstract}
  Applying Large Language Models (LLMs) to graph-structured data usually involves multi-step pipelines in which textual node attributes are compressed into single tokens and further processed by GNNs, discarding most of their semantic content. We introduce the Graph Transformer Language Model (GTLM), which enables a pretrained LLM to process graph topology natively and removes this bottleneck entirely. GTLM injects graph-aware attention biases directly into the LLM's attention modules, adding only 0.015\% structure-related parameters relative to the base model. Training updates only the structural parameters together with a LoRA adapter on the base model. We prove that our bidirectional attention prefix is permutation-equivariant over nodes and that GTLM reduces exactly to the pretrained model when no graph is present. Having no global node ordering, GTLM shows no positional degradation and does not \textit{get lost in the middle}: needle-in-a-graph accuracy stays flat from 1k to 64k tokens and 4× past the training length, while an identically trained flat-text baseline collapses. Comprehensive evaluations show that a GTLM matches or exceeds domain-specific state-of-the-art models on text-attributed graph benchmarks, GraphRAG on WebQSP, and molecular benchmarks, while meaningfully improving over strong baselines on GraphQA. We further show that GTLM's attention heads implicitly learn to simulate message passing, explaining its strength on algorithmic tasks. Together, these results suggest that a minimally adapted pretrained LLM can serve as a general backbone for graph learning.
\end{abstract}

\section{Introduction}

Many real-world problems involve interconnected structures which Large Language Models (LLMs) have to work around due to their inherent inability to perceive structure. The standard way to give such data to an LLM is to serialize the graph into text \citep{fatemi2024talk}. Serialization introduces an arbitrary node order, so the output depends on a choice that carries no meaning.

However, from an architectural point of view, a decoder-only LLM is already a Graph Transformer (GT) \citep{ying2021graphormer}: its input is a path graph over tokens, and the positional encodings under the causal mask are the structure that its attention sees. Moving from a path to any graph should mean changing the way attention sees structure, not adding a separate graph encoder to the model.

Most existing approaches do the latter. Graph-Tokenizing LLMs (GTokenLLMs) \citep{zhang2026graphtokenizinglargelanguagemodels} typically rely on disjointed, multi-step pipelines. These frameworks condense each text-attributed node into a single token via a preliminary encoder, optionally process these representations with Graph Neural Networks (GNNs), and pass the representations into rigid serialization templates. While effective on some benchmarks, this paradigm introduces significant bottlenecks: the fixed input formats constrain versatility, and the lossy compression of rich textual data into solitary tokens prevents the model from accessing fine-grained, node-level information.

\begin{figure}[h]
    \centering
    \includegraphics[width=1\textwidth]{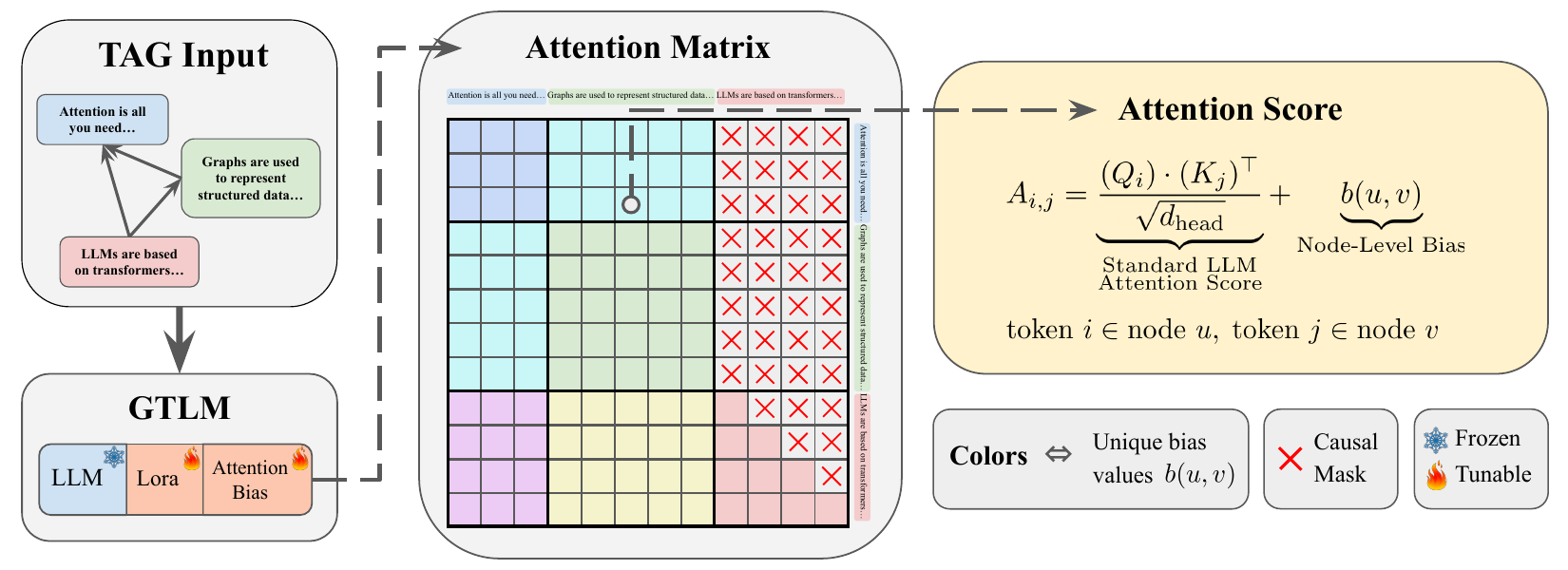}
    \caption{\textbf{GTLM architecture.} Node texts are concatenated, and node-level topological biases $b(u,v)$ are broadcasted to constituent token pairs (where $i \in u, j \in v$) within the attention matrix. Graph structure is integrated using tunable attention biases and LoRA on a frozen base LLM.}
    \label{fig:main_figure}
\end{figure}

In this paper, we introduce the Graph Transformer Language Model (GTLM), a novel architecture that addresses the limitations of multi-step pipelines by integrating structural information directly into the LLM's attention mechanism without lossy text compression (Figure~\ref{fig:main_figure}). Inspired by the Relative Positional Encodings (RPEs) in Graphormer \citep{ying2021graphormer}, GTLM leverages the global topology to broadcast graph attention biases to text token pairs. To preserve permutation equivariance, we flatten the node texts, reset their position IDs, and apply bidirectional prefix attention.

In order to demonstrate the versatility of GTLM, we evaluate it on a wide variety of benchmarks related to text and graphs, all previously attempted by different highly specialized approaches. We empirically show that GTLM is capable of bridging the gap between LLMs and structured data and that the generality of LLMs carries over to these graph-related tasks.

In summary, our main contributions are:

\begin{itemize}
    \item \textbf{Architectural Innovation \& Theoretical Grounding:} We introduce \textbf{GTLM}, a unified architecture that builds upon LLMs and enables them to perceive graphs. Furthermore, we prove that GTLM is a natural generalization of LLMs, satisfying prefix node permutation equivariance and maintaining exact backward compatibility with standard self-attention. 
    \item \textbf{Empirical Performance:} Through extensive evaluation on a wide variety of standardized benchmarks, we demonstrate that GTLM consistently achieves or exceeds state-of-the-art performance against highly specialized niche approaches.
    \item \textbf{Superior Long-Context Retrieval over LLMs:} On a synthetic needle-in-a-graph task, we demonstrate that GTLM does not get \emph{lost in the middle} and that its multi-step retrieval abilities do not degrade with difficulty and context length, while an LLM collapses.
\end{itemize}

To facilitate future research and the adaptation of this architecture to other pretrained LLMs, the code for all model modifications and experiments is made available.\footnote{GTLM GitHub repository: \url{https://github.com/DarioVajda/graph_model}.}

\section{Related Work}
\label{sec:related-work}

\subsection{Large Language Models for Graphs}

Initial approaches to graph-LLM integration relied on explicit text serialization \citep[e.g.,][]{fatemi2024talk}, which inherently suffers from poor graph-aware reasoning and rapid context exhaustion. To mitigate these constraints, research pioneered by \citet{perozzi2024letgraphtalkingencoding} shifted toward multimodal graph alignment, a paradigm further advanced by models such as GraphGPT \citep{tang2024graphgpt}, LLaGA \citep{chen2024llaga}, and RGLM \citep{zhang2026graphtokenizinglargelanguagemodels}. These architectures project graph nodes into ``soft tokens'' within the LLM’s embedding space to facilitate structural processing.

Despite their efficiency, these alignment-based methods introduce a significant \textbf{compressive bottleneck}: rich, multi-sentence node attributes are condensed into solitary dense vectors, abstracting away fine-grained semantics and preventing the LLM from executing token-level attention between nodes. Unlike these multi-step projection frameworks, \textbf{GTLM} eliminates the compression bottleneck entirely by natively routing topological awareness through the attention mechanism while preserving the full text token sequence of every node.

\subsection{Structural Encodings in Graph Transformers}

Since Transformers were adapted to graphs \citep{attention_is_all_you_need, yuan2025survey}, structural injection typically falls into two categories: \textbf{Absolute} Positional Embeddings (APEs), which modify initial node features using structural priors, and \textbf{Relative} Positional Encodings (RPEs), which infer topology via additive biases applied to raw attention scores. Our architecture relies on RPEs, building upon three foundational methods. 

First, \textbf{Shortest Path Distances (SPD)} index a learnable scalar bias based on the discrete shortest path distance between any node pair, natively modulating token interaction strength based on structural proximity without message passing \citep{ying2021graphormer}. Second, \textbf{Relative Random Walk Probabilities (RRWP)} capture complex structural dynamics using $K$-step walk probabilities $\mathbf{P}_{u,v} \in \mathbb{R}^K$ derived from the graph's transition matrix \citep{ma2023graphinductivebiasestransformers}. 

Finally, to capture directed graph topology, the \textbf{Magnetic Laplacian} utilizes a matrix that encodes edge directionality into a phase matrix $\mathbf{\Theta}^{(q)} \in \mathbb{R}^{N \times N}$ to obtain the Normalized Magnetic Laplacian $\mathbf{L}_N^{(q)} \in \mathbb{C}^{N \times N}$ \citep{magnet_citation_zhang}. By performing spectral decomposition on $\mathbf{L}_N^{(q)}$, basis-invariant edge-level features $\mathbf{K} \in \mathbb{C}^{N \times N \times d_{\text{Mag}}}$ can be constructed, similar to \citet{huang2025what}.

\section{Methodology}
To unify the graph and text modalities, we represent all data through an underlying graph structure where nodes (and optionally edges) consist of text sequences. The entire input graph is processed by a modified LLM, inspired by GTs, which we call \textbf{Graph Transformer Language Model (GTLM)}. It accounts for both the global topology and the relative positions of text tokens within each node.

\subsection{Problem setup}

Let $\mathcal{G} = (\mathcal{V}, \mathcal{E}, \mathcal{T})$ be a Text-Attributed Graph (TAG), with vertices $\mathcal{V}=\{v_0, v_1, \dots, v_N\}$, edges $\mathcal{E} \subseteq \mathcal{V} \times \mathcal{V}$, and tokenized textual attributes $\mathcal{T}=\{t_0, t_1, \dots, t_N\}$. Let $v_\text{target}=v_0$ denote the \textbf{target node} for which we want to perform autoregressive text generation, with its corresponding text denoted as $y = t_0$. Further, let $L$ be the total sequence length equal to $\sum_{i=0}^{N}\text{len}(t_i)$.

Assuming all context nodes $\{v_1, \dots, v_N\}$ are permutation invariant, we define an arbitrary permutation $\pi \in S_N$. Finally, let $S$ be a set of \textbf{structural features} derived from $(\mathcal{V}, \mathcal{E})$, which GTLM utilizes for attention bias computation. Our objective is to maximize the log-likelihood of the target sequence $y$ given the context nodes and the structural features. The autoregressive generation of token $y_j$, conditioned on the target prefix $y_{<j}$ and the permuted context tokens, is formulated as:
$
\max_{\theta} \quad \sum_{j=1}^{|y|} \log P_{\theta} \Big( y_j \mid \big[ t_{\pi(1)} \oplus \dots \oplus t_{\pi(N)} \oplus y_{<j} \big], S \Big)
$
, where $P_\theta$ denotes the probability distribution parameterized by the GTLM network $\theta$, and $\oplus$ denotes sequence concatenation.

\subsection{Modified Attention}
\label{sec:modified-attention}

The main innovation of our approach is the combination of sequential RPEs traditionally used by LLMs, such as RoPE \citep{su2023roformerenhancedtransformerrotary}, and RPEs for graphs. To ensure \textbf{permutation equivariance}, the position IDs for RoPE are counted from zero onwards inside of each node individually. While in this paper we primarily focus on RoPE, our architecture is compatible with other PEs such as ALiBi \citep{press2022train} or dual RoPE \citep{gemma3report2025}, which we empirically demonstrate in Appendix \ref{app:other-pos-encodings}.

The raw attention score $A^{(l, h)}_{i,j}$ between tokens $i$ and $j$ (inside of nodes $u$ and $v$, respectively) for the $h$-th attention head within the $l$-th layer of GTLM is calculated as shown below:

\[
A^{(l, h)}_{i,j} = 
\underbrace{\frac{(Q^{(l, h)}_i) \cdot (K^{(l, h)}_j)^\top}{\sqrt{d_{\text{head}}}}}_{\text{Standard LLM Attention Score}} + 
\underbrace{b_{\text{SPD}}^{(l,h)}(u,v)}_{\text{SPD Bias}} + 
\underbrace{b_{\text{RRWP}}^{(l,h)}(u,v)}_{\text{RRWP Bias}} + 
\underbrace{b_{\text{Mag}}^{(l,h)}(u,v)}_{\text{Magnetic Laplacian Bias}}.
\]

Critically, all nodes before the target node are treated as a prefix, following the PrefixLM architecture \citep{prefixLM_citation_raffel}, where a fully-visible (non-causal) mask is applied to the prefix to allow for bidirectional context, while maintaining causal constraints for the target node's text generation. In our architecture, we propose three main ways of encoding the structural relative position of tokens and we can utilize any combination of the following approaches simultaneously, each relying on mathematical features derived in Appendix~\ref{app:structural_derivations}.

\textbf{Shortest Path Distance (SPD) Bias.} This is a learned lookup table mapping the SPD of the corresponding nodes to an attention bias added to the raw attention scores.
\[b_{\text{SPD}}^{(l,h)}(u,v) =\text{b}^{(l,h)}_{\text{spd}}(D_{\text{hops}}(u,v)), \text{ where } b_{\text{spd}} \in \mathbb{R}^{L \times H \times \text{max\_spd}} \text{ is learned, with } b_{\text{SPD}}^{(l,h)}(u,u)=0.\]

\textbf{Relative Random Walk Probabilities (RRWP) Bias.} Building upon the structural formulation introduced earlier, let $\mathbf{P}_{u,v} \in \mathbb{R}^K$ represent the $K$-step random walk probabilities between nodes $u$ and $v$. We transform these initial node-pair-level features into a scalar attention bias using a layer-specific, learned Multi-Layer Perceptron ($\text{MLP}_{\text{RRWP}}$). Crucially, to preserve the intra-node attention properties essential for compatibility with pretrained LLMs, this bias is explicitly zeroed out for tokens belonging to the same node:
\[
b_{\text{RRWP}}^{(l,h)}(u,v) = \begin{cases}
    0 & u=v \\
    [\text{MLP}_{\text{RRWP}}^{(l)}(\mathbf{P}_{u,v})]_h & u\neq v
\end{cases}
\space ,
\]
where $\text{MLP}_{\text{RRWP}}^{(l)} : \mathbb{R}^K \rightarrow \mathbb{R}^{\text{n\_heads}}$ maps the structural features to the respective attention heads of the $l$-th layer.

\textbf{Magnetic Laplacian Bias.} To natively incorporate directed spectral geometry into our model, we derive relative attention biases from the basis-invariant Magnetic Laplacian kernel $\mathbf{K} = \mathbf{V} \text{diag}(\phi(\lambda)) \mathbf{V}^\dagger \in \mathbb{C}^{N \times N \times d_{\text{Mag}}}$, as introduced in Section \ref{sec:related-work}. To instantiate the required permutation-invariant transformation for the eigenvalues, we specifically employ a Deep Set architecture \citep{zaheer2017deep}:
\[ \phi(\lambda_i) = \sigma(\mathbf{W}_2 [\text{Linear}(\lambda_i) \oplus \text{mean}(\text{Linear}(\lambda))]) \in \mathbb{R}^{d_\text{Mag}}, \]
where $d_\text{Mag}$ is a hyperparameter dictating the spectral feature dimension. By decomposing the resulting node-pair interactions into their real and imaginary components, $\text{Re}(\mathbf{K})$ and $\text{Im}(\mathbf{K})$, we map the rich spectral information to a scalar bias for each attention head via a learned MLP. 

Consistent with our approach of maintaining intra-node attention integrity for compatibility with pretrained LLMs, we explicitly zero out this bias when tokens belong to the same node:
\[
b_{\text{Mag}}^{(l,h)}(u,v) = \begin{cases} 
0 & u=v \\
[\text{MLP}_{\text{Mag}}^{(l)} ( [ \text{Re}(\mathbf{K}_{u,v}) \oplus \text{Im}(\mathbf{K}_{u,v}) ] )]_h & u \neq v
\end{cases}
\space .
\]

\subsection{Theoretical Properties of GTLM}
\label{sec:theoretical-properties}

Our proposed GTLM architecture satisfies two fundamental properties that successfully bridge the theoretical gap between Graph Transformers and Large Language Models. We provide proofs for both theoretical property claims, as well as empirical results to verify them in Appendix~\ref{app:properties-proof}.

\textbf{Property 1---Prefix Node Permutation Equivariance:} The internal representations and final outputs of GTLM are equivariant with respect to the serialization order of the prefix nodes.

\textit{Explanation:} When flattening the nodes of a graph into a sequence, their order does not affect the attention scores between any two tokens, because the attention score depends only on the position ID (which is independent of the node order, due to the resetting of the RoPE ID) and the relative bias at the node level, which is by definition invariant under permutations.

\textbf{Property 2---Pretraining Backward Compatibility:} For any graph $\mathcal{G}$ consisting of a single node (representing a standard text sequence), the GTLM forward pass is mathematically identical to that of the unmodified base LLM.

\textit{Explanation:} Since the condition that the bias is zero on the diagonal of the node-level bias matrix is imposed by definition, the bias is a null operation in a single-node graph, as is resetting the RoPE IDs. Finally, the full input is causally masked, since the only node is the target node---making the architecture mathematically equivalent to the base LLM.

\section{Experiments}
\label{sec:experiments}

We conduct comprehensive experiments across a diverse set of established benchmarks. Furthermore, we conduct a thorough ablation study (Section~\ref{sec:ablation}) evaluating the importance of each attention bias type. Unless stated otherwise, \texttt{Llama-3.2-1B} \citep{llama3herdofmodels} served as the base model, and a Low-Rank Adapter (LoRA) \citep{hu2021loralowrankadaptationlarge} was used. Specific details about the training setup and hyperparameters for each experiment can be found in Appendices~\ref{app:experimental-details}--\ref{app:context-experiment}. Each result that reports standard deviations was measured using three random seeds; otherwise only one was used. Our base-model scaling analysis (Appendix \ref{app:scaling-base}) finds that performance is mostly saturated at 1B for classification tasks, while open-ended generation (such as WebQSP, CWQ, and ChEBI-20) benefits from scaling to an 8B-parameter base model.

\subsection{GraphQA}
\label{sec:graphqa}

GraphQA is a question-answering benchmark about abstract graphs and their properties (node degree, reachability, etc.). This problem was previously attempted by pure LLMs \citep{fatemi2024talk}, and later by the GraphToken architecture \citep{perozzi2024letgraphtalkingencoding}. While GraphToken's performance meaningfully depended on the graph preprocessing method, our GTLM demonstrated exceptional performance and surpassed the baselines in all but one task, as seen in Table~\ref{tab:graphqa-results}. The experiment was conducted under two settings: (1) where the input graphs were unchanged (\textit{Standard}), and (2) where the input graphs were transformed into \textit{Incidence} graphs---a bipartite transformation where nodes represent the original vertices and edges.

\begin{table}[ht]
    \caption{GraphQA per-task accuracy ($\%$) (\textbf{best}, \underline{second-best}). The baselines are a Zero-Shot LLM and the top-3 highest scoring graph encoding methods used by GraphToken.}
    \label{tab:graphqa-results}
    \vspace{0.5em}
    \centering
    \renewcommand{\arraystretch}{1.1}
    
    \setlength{\tabcolsep}{9pt} 
    
    \renewcommand{\llnote}[1]{\rlap{\textsubscript{\tiny~#1}}}
    \renewcommand{\std}[1]{\rlap{~$\pm$#1}}
    
    \resizebox{\textwidth}{!}{
    \begin{tabular}{@{}l *{4}{c} *{2}{c} *{3}{c}@{\hspace{1.5em}}}
        \toprule
        & \multicolumn{4}{c}{Graph Tasks} & \multicolumn{2}{c}{Node Tasks} & \multicolumn{3}{c}{Edge Tasks} \\
        \cmidrule(lr){2-5} \cmidrule(lr){6-7} \cmidrule(lr){8-10}
        & Node & Edge & Cycle & Triangle & Node & Conn. & Reach- & Edge & Shortest \\
        Method & Count & Count & Check & Count & Degree & Nodes & ability & Exist. & Path \\
        \midrule
        Zero-Shot LLM       & 21.7 & 12.4 & 76.0 & 1.5 & 14.0 & 14.7 & 84.9 & 44.5 & 11.5 \\
        3rd GraphToken      & 79.2\llnote{MPNN} & 26.4\llnote{MHA} & 96.2\llnote{MHA} & 23.4\llnote{HGT} & 26.6\llnote{HGT} & 24.4\llnote{MHA} & 93.2\llnote{MHA} & 68.0\llnote{GCN} & 60.4\llnote{GCN} \\
        2nd GraphToken      & 91.2\llnote{MHA} & 36.8\llnote{MPNN} & 96.4\llnote{GCN} & 26.6\llnote{MHA} & 55.2\llnote{MHA} & 25.0\llnote{MPNN} & 94.2\llnote{NS} & 71.8\llnote{HGT} & 60.8\llnote{MHA} \\
        1st GraphToken      & \underline{99.6}\llnote{NS} & 42.6\llnote{ES} & 96.4\llnote{ES} & \textbf{34.8}\llnote{MPNN} & \underline{96.2}\llnote{MPNN} & 26.4\llnote{GCN} & 94.4\llnote{HGT} & 73.8\llnote{MHA} & 63.8\llnote{MPNN} \\
        \midrule
        \textbf{GTLM} {\scriptsize(Incidence)} & 99.1\sstd{0.1} & \textbf{62.2}\sstd{2.0} & \textbf{98.7}\sstd{0.3} & \underline{31.6}\sstd{0.5} & 91.8\sstd{1.0} & \textbf{98.8}\sstd{0.3} & \underline{98.7}\sstd{0.1} & \underline{99.6}\sstd{0.2} & \textbf{91.7}\sstd{1.8} \\
        \textbf{GTLM} {\scriptsize(Standard)}  & \textbf{100}\sstd{0.0} & \underline{56.9}\sstd{1.7} & \underline{96.7}\sstd{0.6} & 30.9\sstd{0.6} & \textbf{99.7}\sstd{0.1} & \underline{90.7}\sstd{0.3} & \textbf{99.0}\sstd{0.0} & \textbf{99.8}\sstd{0.2} & \underline{90.1}\sstd{1.7} \\
        \bottomrule
    \end{tabular}
    }
\end{table}

\subsection{Text-Attributed Graph Benchmarks}
\label{sec:tag_benchmarks}
The second suite of tests we evaluate GTLM with focuses on node classification in Text-Attributed Graphs. We evaluate it on the following four benchmarks: \textit{Cora}, \textit{Pubmed} \citep{yang2016revisiting}, and \textit{OGBN-Arxiv} \citep{hu2020open} for citation networks, and \textit{Reddit} \citep{huang2024can} for a social network. We use Accuracy ($\uparrow$) and F1 score ($\uparrow$) to evaluate GTLM, as those are the widely adopted standards for these benchmarks. To establish a robust evaluation framework, we benchmark GTLM against the suite of baselines curated by \citet{zhang2026graphtokenizinglargelanguagemodels} with matching data splits. We categorize these baselines into two primary groups: \textit{Graph-native architectures} and \textit{Multimodal LLMs}.

\begin{table}[ht]
\centering
\caption{Text-Attributed Graph node classification results (\textbf{best}, \underline{second-best}).}
\label{tab:node_classification}
\small
\setlength{\tabcolsep}{8pt}
\resizebox{\columnwidth}{!}{%
\begin{tabular}{@{}l cc cc cc cc@{}}
\toprule
 & \multicolumn{2}{c}{Cora} & \multicolumn{2}{c}{Pubmed} & \multicolumn{2}{c}{OGBN-Arxiv} & \multicolumn{2}{c}{Reddit} \\
\cmidrule(lr){2-3}\cmidrule(lr){4-5}\cmidrule(lr){6-7}\cmidrule(lr){8-9}
Method & Acc & F1 & Acc & F1 & Acc & F1 & Acc & F1 \\
\midrule
\multicolumn{9}{@{}l}{\textit{Graph-native architectures}} \\
GCN \citep{kipf2016semi} & 89.30 & 87.95 & 89.48 & 89.04 & 73.83 & 54.97 & 63.19 & 62.49 \\
GraphSAGE \citep{hamilton2017inductive} & 87.82 & 86.77 & 90.54 & 90.51 & 74.47 & 56.41 & 58.51 & 58.41 \\
GAT \citep{velickovic2018graph} & 87.08 & 86.32 & 88.11 & 87.60 & 74.39 & 57.71 & 61.78 & 65.38 \\
NodeFormer \citep{wu2022nodeformer} & 87.30 & 85.03 & 89.92 & 89.63 & 68.69 & 50.64 & 68.06 & 67.89 \\
GRACE \citep{zhu_deep_graph_crl2020} & 89.11 & 88.12 & 86.89 & 86.58 & 72.00 & 50.79 & 65.39 & 65.39 \\
\midrule
\multicolumn{9}{@{}l}{\textit{Multimodal LLMs}} \\
GraphText-ICL \citep{zhao2024graphtext} & 82.29 & 81.16 & 56.72 & 52.55 & 68.67 & 51.67 & 56.50 & 52.23 \\
GraphText-SFT \citep{zhao2024graphtext} & 84.69 & 83.77 & 87.80 & 87.68 & 49.88 & 19.46 & 62.20 & 62.07 \\
GraphGPT \citep{tang2024graphgpt} & 83.92 & 82.72 & 81.76 & 80.99 & 71.59 & 54.58 & 62.11 & 62.39 \\
LLaGA \citep{chen2024llaga} & 88.75 & 87.87 & 90.34 & 90.25 & 74.61 & 56.48 & 66.38 & 66.38 \\
RGLM-Decoder \citep{zhang2026graphtokenizinglargelanguagemodels} & 89.85 & \underline{89.39} & 91.15 & 91.07 & 75.00 & 56.92 & \textbf{68.64} & \textbf{68.54} \\
RGLM-Similarizer \citep{zhang2026graphtokenizinglargelanguagemodels} & 89.67 & 88.96 & \underline{91.61} & \underline{91.51} & \underline{75.14} & \underline{57.74} & 66.79 & 66.77 \\
RGLM-Denoiser \citep{zhang2026graphtokenizinglargelanguagemodels} & \textbf{90.22} & 89.35 & 90.95 & 90.90 & 75.10 & 56.84 & 67.64 & 67.50 \\
\midrule
\multirow{3}{*}{\begin{tabular}{@{}l@{}}\textbf{GTLM}\\{\scriptsize (Llama-3.2-1B)}\end{tabular}} \hspace{4em} \textit{Best} & \underline{90.04} & \textbf{92.14} & \textbf{94.70} & \textbf{94.84} & \textbf{76.53} & \textbf{70.02} & \underline{68.09} & \underline{68.00} \\
\hphantom{\begin{tabular}{@{}l@{}}\textbf{GTLM}\\{\scriptsize (Llama-3.2-1B)}\end{tabular}} \hspace{4em} \textit{Mean} & 88.99 & 91.31 & 94.57 & 94.76 & 76.07 & 69.40 & 67.80 & 67.71 \\
\hphantom{\begin{tabular}{@{}l@{}}\textbf{GTLM}\\{\scriptsize (Llama-3.2-1B)}\end{tabular}} \hspace{4em} \textit{Std} & 1.12 & 0.90 & 0.11 & 0.10 & 0.54 & 0.86 & 0.37 & 0.32 \\
\bottomrule
\end{tabular}%
}
\end{table}

As detailed in Table~\ref{tab:node_classification}, GTLM consistently achieves state-of-the-art or highly competitive performance across all four benchmarks. The model surpasses all baselines on \textit{Pubmed} and \textit{OGBN-Arxiv} across both accuracy and F1 score. On \textit{Cora}, GTLM establishes a new best F1 score while maintaining top-tier accuracy, and it remains a strong runner-up on the \textit{Reddit} dataset.

\subsection{GraphRAG Evaluation}
\label{sec:graphrag}
Knowledge-graph question answering is highly relevant to industrial applications. This area has seen advancements in highly specific methods and the current state of the art is established by agentic approaches, such as \textbf{GraphWalker}~\citep{graphwalker2026} and \textbf{TRACE}~\citep{trace2026}, which lead on both benchmarks by querying the graph interactively, so no single subgraph bounds them. Therefore, we leave them out of the headline comparison, as our aim is to focus on graph comprehension, rather than optimizing the subgraph retrieval method. We compare against systems that, like ours, extract once under a fixed budget---\textbf{RoG}~\citep{rog2024}, \textbf{GNN-RAG} \citep{gnnrag2025} with its retrieval-augmented variant \textbf{GNN-RAG+RA} and the variant reading the exact same SR subgraphs we do~\citep{sr2022}, and \textbf{SubgraphRAG}~\citep{subgraphrag2025}---on \textbf{WebQSP}~\citep{webqsp2016} and \textbf{CWQ}~\citep{cwq2018}.

\begin{table}[ht]
\centering
\caption{Knowledge graph QA benchmarks WebQSP and CWQ (\textbf{best}, \underline{second-best}).}
\label{tab:kgqa-main}
\small
\setlength{\tabcolsep}{12pt}
\begin{tabular}{@{}ll ccc ccc@{}}
\toprule
 & & \multicolumn{3}{c}{WebQSP} & \multicolumn{3}{c}{CWQ} \\
\cmidrule(lr){3-5}\cmidrule(lr){6-8}
Method & Reader & Hit & H@1 & F1 & Hit & H@1 & F1 \\
\midrule
\multicolumn{8}{@{}l}{\textit{Own retrieval}} \\
RoG & Llama-2-7B & 85.7 & 80.0 & 70.8 & 62.6 & 57.8 & 56.2 \\
GNN-RAG & Llama-2-7B & 85.7 & 80.6 & 71.3 & 66.8 & 61.7 & 59.4 \\
GNN-RAG + RA & Llama-2-7B & \textbf{90.7} & \textbf{82.8} & 73.5 & \textbf{68.7} & \underline{62.8} & \underline{60.4} \\
SubgraphRAG & GPT-4o-mini & \underline{90.1} & -- & \underline{77.5} & 62.0 & -- & 54.1 \\
\midrule
\multicolumn{8}{@{}l}{\textit{SR-retrieved (matched inputs)}} \\
GNN-RAG (SR) & Llama-2-7B & 83.4 & 78.9 & 69.8 & 60.6 & 55.6 & 53.3 \\
\textbf{GTLM (ours)} & Llama-3.1-8B & 85.9 & \underline{82.5} & \textbf{77.9} & \underline{67.3} & \textbf{64.5} & \textbf{62.1} \\
\bottomrule
\end{tabular}
\end{table}

On matched inputs GTLM improves on GNN-RAG's reasoner-plus-reader pipeline by 8.1 F1 on WebQSP (77.9 vs.\ 69.8) and 8.8 on CWQ (62.1 vs. 53.3). SR's subgraphs are far smaller than the baselines' two-hop pools (median 29 nodes vs.\ 1530 on WebQSP, 94 vs.\ 1481 on CWQ). On WebQSP the coverage costs 5.2 points of ceiling (89.1 vs.\ 94.3) and GTLM overtakes GNN-RAG on its own retrieval regardless (77.9 vs.\ 71.3); on CWQ the ceilings coincide (79.0 vs.\ 79.5), so GTLM's 2.7-point F1 lead over GNN-RAG (62.1 vs. 59.4) reflects stronger reasoning. Appendix \ref{app:graphrag} reports the chosen hyperparameters and the coverage ceiling measurements.

\subsection{Molecule Benchmarks}
\label{sec:molecules}

Molecule processing tasks have been attempted by approaches centered around LLMs. These methods include those using purely textual SMILES encodings \citep{runcie2026molecularrepresentationslargelanguage}, and others which take an approach more similar to ours, where they modify the LLM's architecture to directly process the molecule graphs. Following the evaluation setup of \citet{cao2025instructmol}, we compare GTLM to other existing methods on \textbf{MoleculeNet} \citep{wu2018moleculenet} property prediction (Table \ref{tab:molecules-prop-pred}) and \textbf{ChEBI-20} \citep{edwards2021text2mol} molecule captioning (Table \ref{tab:molecules-captioning}).

\begin{table}[ht]
\centering
\caption{MoleculeNet property prediction, \textbf{ROC-AUC} ($\uparrow$). \textbf{Best} and \underline{second-best} marked among LLM-based methods.}
\label{tab:molecules-prop-pred}
\small
\setlength{\tabcolsep}{12pt}
\begin{tabular}{@{}ll ccc@{\hphantom{\scriptsize$\,\pm\,$11}}}
\toprule
Method & Input & BACE & BBBP & HIV \\
\midrule
\multicolumn{5}{@{}l}{\textit{Chemistry-pretrained specialists}} \\
Uni-Mol \citep{zhou2023unimol} & 3D conformers & 85.7 & 72.9 & 80.8 \\
MolCA \citep{liu2023molca} & SMILES + graph & 79.8 & 70.0 & -- \\
GraphCL \citep{you2020graphcl} & graph & 75.3 & 69.7 & 78.5 \\
ChemBERTa-2 \citep{ahmad2022chemberta2} & SMILES & 73.5 & 69.8 & 79.3 \\
\midrule
\multicolumn{5}{@{}l}{\textit{Fine-tuned LLM-based methods}} \\
InstructMol-GS \citep{cao2025instructmol} & graph + SMILES & \underline{82.1} & \textbf{72.4} & 68.9 \\
InstructMol-G  \citep{cao2025instructmol} & graph & \textbf{84.3} & 68.6 & \underline{74.0} \\
Llama-2-7B-chat \citep{touvron2023llama2} & SMILES & 74.8 & 65.6 & 62.3 \\
Vicuna-v1.3-7B \citep{chiang2023vicuna} & SMILES & 68.3 & 60.1 & 58.1 \\
\textbf{GTLM} \scriptsize{(Llama-3.2-1B)} & graph & 82.0\sstd{1.2} & \underline{70.6}\sstd{2.3} & \textbf{76.9}\sstd{1.7} \\
\bottomrule
\end{tabular}
\end{table}

\begin{table}[ht]
\centering
\caption{ChEBI-20 molecule captioning. \textbf{Best} and \underline{second-best} marked among LLM-based methods.}
\label{tab:molecules-captioning}
\small
\setlength{\tabcolsep}{8pt}
\resizebox{\columnwidth}{!}{%
\begin{tabular}{@{}ll cccc@{\hspace{1.7em}}}
\toprule
Method & Input & BLEU-2 & BLEU-4 & ROUGE-L & METEOR \\
\midrule
\multicolumn{6}{@{}l}{\textit{Chemistry-pretrained sequence models}} \\
MolT5-Large \citep{edwards2022molt5} & SMILES & 0.594 & 0.508 & 0.594 & 0.614 \\
Text+Chem T5 \citep{christofidellis2023textchemt5} & SMILES & 0.625 & 0.542 & 0.622 & 0.648 \\
MolCA \citep{liu2023molca} & SMILES + gr\rlap{aph} & 0.620 & 0.531 & 0.618 & 0.651 \\
\midrule
\multicolumn{6}{@{}l}{\textit{Fine-tuned LLM-based methods}} \\
Mol-Instructions \citep{fang2024molinstructions} & SELFIES & 0.249 & 0.171 & 0.289 & 0.271 \\
BioMedGPT-10B \citep{luo2023biomedgpt} & graph & 0.234 & 0.141 & 0.332 & 0.308 \\
InstructMol-GS \citep{cao2025instructmol} & graph + \rlap{SMILES} & 0.475 & 0.371 & 0.502 & 0.509 \\
InstructMol-G \citep{cao2025instructmol} & graph & 0.466 & 0.365 & 0.479 & 0.491 \\
\textbf{GTLM} \scriptsize{(Llama-3.2-1B)} & graph & \underline{0.534}\sstd{0.009} & \underline{0.437}\sstd{0.011} & \underline{0.551}\sstd{0.006} & \underline{0.567}\sstd{0.008} \\
\textbf{GTLM} \scriptsize{(Llama-3.1-8B)} & graph & \textbf{0.572}\sstd{0.003} & \textbf{0.482}\sstd{0.003} & \textbf{0.583}\sstd{0.002} & \textbf{0.602}\sstd{0.002} \\
\bottomrule
\end{tabular}%
}
\end{table}

We find that GTLM significantly outperforms general-purpose LoRA-tuned LLMs, and performs on par with the 7B InstructMol models on property prediction, despite their chemistry-pretrained GNN encoder. Finally, GTLM demonstrates strong performance on ChEBI-20 molecule captioning, with a $0.097$ BLEU-2 advantage over the best LLM-based method InstructMol-GS ($0.572$ vs $0.475$), nearly matching chemistry-pretrained specialist models. Additional details about these experiments are provided in Appendix \ref{app:molecule-benchmarks}.

\subsection{Long-Context Retrieval}
\label{sec:long-context}

Finally, we compare GTLM against a standard LLM under a matching training budget on a synthetic \textbf{needle-in-a-graph} task, where the model must follow a $k$-hop path in a graph to retrieve the answer (Table~\ref{tab:nt-scaling}). As graph size $N$ and per-node token count $T$ grow, GTLM maintains high accuracy. The LLM collapses to $8.5\%$ at $(N{=}128, T{=}512)$ on $4$-hop questions compared to GTLM's $98.0\%$.

This gap is a result of two properties. First, by \textbf{Property 1} (Prefix Node Permutation Equivariance), GTLM is invariant to node order and cannot get \emph{lost in the middle}. On the other hand, the LLM suffers up to 57 pp fluctuations across node permutations (documented in Appendix~\ref{app:context-experiment}). Second, since GTLM resets RoPE indices for all nodes individually, token positions scale with node length $T$ rather than total input sequence length $N \times T$, enabling seamless out-of-distribution (OOD) context length generalization ($*$ cells) where standard LLMs struggle.

\begin{table}[ht]
\centering
\caption{\textbf{Needle-in-a-graph performance} ($\%$) across graph size $N$ and token count per node $T$ for $k$-hop questions ($k \in \{2, 4\}$). \textbf{*} marks cells outside of the training distribution.}
\label{tab:nt-scaling}
\small
\setlength{\tabcolsep}{5pt}
\newcommand{\s}{\rlap{*}}
\resizebox{\textwidth}{!}{
\begin{tabular}{@{}r@{\hspace{1.4em}} cccc@{\hspace{1.6em}} cccc@{\hspace{2.2em}} cccc@{\hspace{1.6em}} cccc@{\hspace{0.6em}}}
\toprule
 & \multicolumn{8}{c@{\hspace{2.2em}}}{$k = 2$} & \multicolumn{8}{c@{\hspace{0.6em}}}{$k = 4$} \\
\cmidrule(r{2.2em}){2-9} \cmidrule(r{0.6em}){10-17}
 & \multicolumn{4}{c@{\hspace{1.6em}}}{GTLM} & \multicolumn{4}{c@{\hspace{2.2em}}}{LLM} & \multicolumn{4}{c@{\hspace{1.6em}}}{GTLM} & \multicolumn{4}{c@{\hspace{0.6em}}}{LLM} \\
\cmidrule(r{1.6em}){2-5} \cmidrule(r{2.2em}){6-9} \cmidrule(r{1.6em}){10-13} \cmidrule(r{0.6em}){14-17}
$N \setminus T$ & 64 & 128 & 256 & 512 & 64 & 128 & 256 & 512 & 64 & 128 & 256 & 512 & 64 & 128 & 256 & 512 \\
\midrule
16  & 100 & 100 & 100 & 100    & 95.5 & 94.0 & 95.0 & 95.5    & 92.5 & 91.0 & 93.5 & 89.0    & 27.5 & 31.0 & 35.0 & 31.0 \\
32  & 100 & 100 & 100 & 100    & 94.0 & 93.0 & 92.0 & 91.0    & 97.0 & 95.0 & 90.5 & 94.0    & 17.0 & 20.0 & 19.0 & 20.0 \\
64  & 100 & 100 & 100 & 100\s  & 90.5 & 90.5 & 87.5 & 87.0\s  & 95.5 & 95.5 & 95.0 & 94.0\s  & 12.0 & 16.0 & 12.0 & 11.5\s \\
128 & 100 & 100 & 100\s & 100\s & 88.0 & 83.0 & 84.0\s & 78.0\s & 98.0 & 96.5 & 97.5\s & 98.0\s & 8.5  & 10.5 & 6.5\s  & 8.5\s  \\
\bottomrule
\end{tabular}
}
\end{table}

\subsection{Ablation Study}
\label{sec:ablation}
To justify the simultaneous use of three types of attention biases, we conduct a thorough ablation study on the GraphQA benchmark, which was chosen due to its abstract nature, where the models are evaluated purely on their ability to process and reason over graph topologies, rather than textual information. Furthermore, we chose to represent the graphs in their standard format, as it isolates GTLM's ability to infer connectivity information solely from the attention biases.

\begin{table}[ht]
    \caption{\textbf{Ablation study (GraphQA)} results, comparing the accuracies ($\%$) of standard GTLM to its variants without each attention bias type individually.}
    \label{tab:ablation-study}
    \vspace{0.5em}
    \centering
    \renewcommand{\arraystretch}{1.1}
    
    \setlength{\tabcolsep}{9pt} 
        
    \resizebox{\textwidth}{!}{
    \begin{tabular}{@{}l *{4}{c} *{2}{c} *{3}{c}@{\hspace{1.5em}}}
        \toprule
        & \multicolumn{4}{c}{Graph Tasks} & \multicolumn{2}{c}{Node Tasks} & \multicolumn{3}{c}{Edge Tasks} \\
        \cmidrule(lr){2-5} \cmidrule(lr){6-7} \cmidrule(lr){8-10}
        & Node & Edge & Cycle & Triangle & Node & Conn. & Reach- & Edge & Shortest \\
        Method & Count & Count & Check & Count & Degree & Nodes & ability & Exist. & Path \\
        \midrule
        Standard     & 99.7\sstd{0.5} & 57.2\sstd{2.7} & 97.7\sstd{0.8} & 31.5\sstd{0.8} & 99.9\sstd{0.1} & 94.3\sstd{0.1} & 98.4\sstd{0.0} & 99.9\sstd{0.1} & 92.7\sstd{1.7} \\
        \midrule
        w/o SPD      & 99.9\sstd{0.1} & 59.3\sstd{5.1} & 96.8\sstd{0.2} & 31.5\sstd{0.8} & 99.9\sstd{0.1} & 93.2\sstd{0.7} & 98.5\sstd{0.1} & 99.9\sstd{0.2} & 93.5\sstd{0.5} \\
        w/o RRWP     & 100\sstd{0.0}  & 54.4\sstd{4.5} & 96.9\sstd{0.2} & 30.7\sstd{1.1} & 99.9\sstd{0.2} & 93.3\sstd{0.4} & 97.9\sstd{0.1} & 99.6\sstd{0.2} & 91.9\sstd{0.6} \\
        w/o Magnetic & 100\sstd{0.0}  & 8.9\sstd{0.5}  & 94.5\sstd{0.4} & 21.0\sstd{2.1} & 99.9\sstd{0.1} & 94.4\sstd{1.0} & 97.5\sstd{0.2} & 99.3\sstd{0.3} & 87.5\sstd{3.9} \\
        \bottomrule
    \end{tabular}
    }
\end{table}

As seen in Table~\ref{tab:ablation-study}, we find that the Magnetic Bias is the most crucial for strong performance, RRWP demonstrated more modest gains only in some specific tasks, and SPD's effect is no greater than the per-seed noise. However, since the SPD bias does not introduce many new parameters or any computational overhead, we still opted to use all three bias types across our other experiments. Additionally, in Appendix \ref{app:performance-attribution}, we show that the performance gains can be attributed to both: (1) fixing the compression bottleneck and (2) the graph-derived attention biases.

\section{Discussion}
\label{sec:discussion}

In this section, we look beyond the raw performance metrics to analyze the underlying mechanisms and practical properties of GTLM. We demonstrate GTLM's ability to simulate message passing, and discuss its parameter efficiency, implementation, and future research directions.

\subsubsection*{Simulated Message Passing}
\label{sec:simulated-message-passing}
\begin{figure}[ht]
    \centering
    \begin{subfigure}[b]{0.44\textwidth}
        \centering
        \includegraphics[width=1\textwidth]{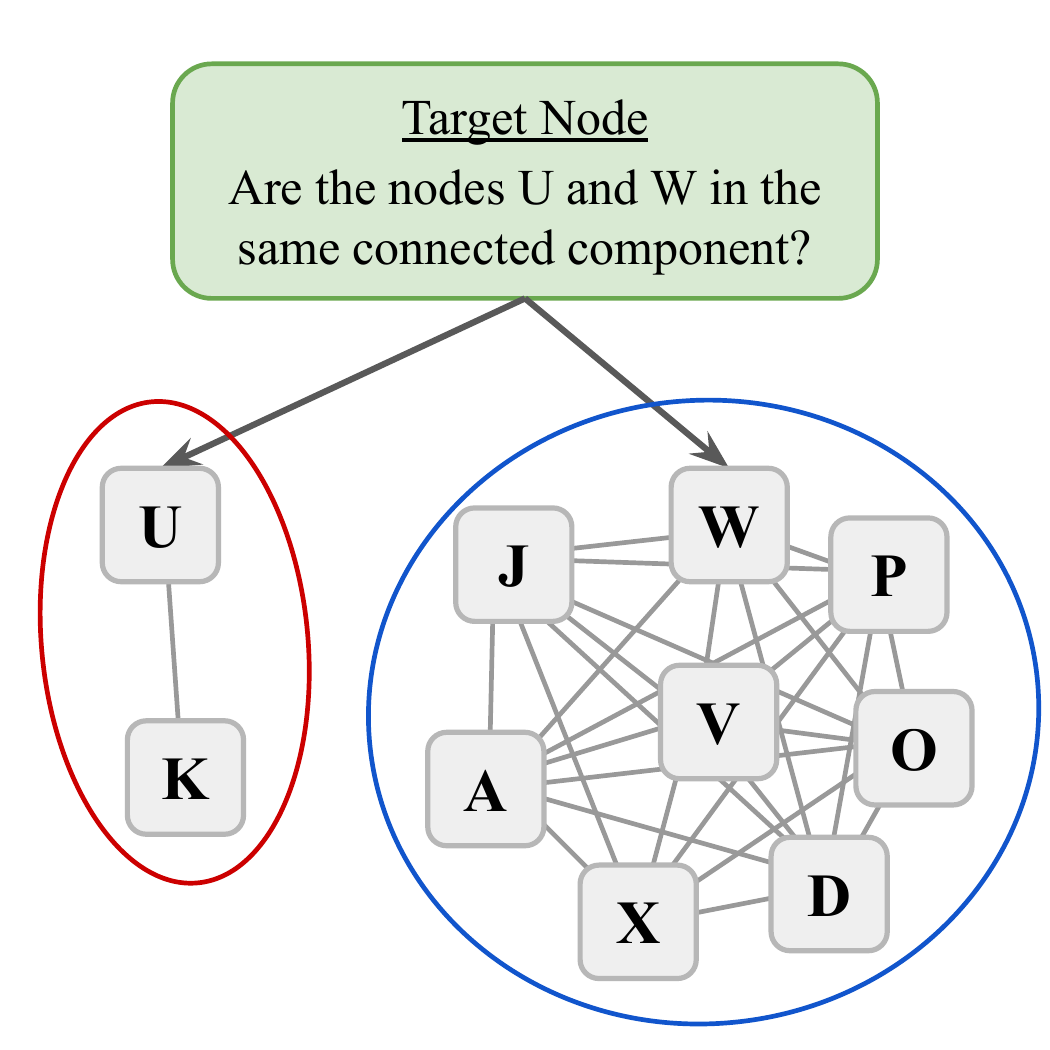}
        \caption{Visualization of the synthetic clique-detection task.}
        \label{fig:clique_task}
    \end{subfigure}
    \hfill
    \begin{subfigure}[b]{0.44\textwidth}
        \centering
        \includegraphics[width=1\textwidth]{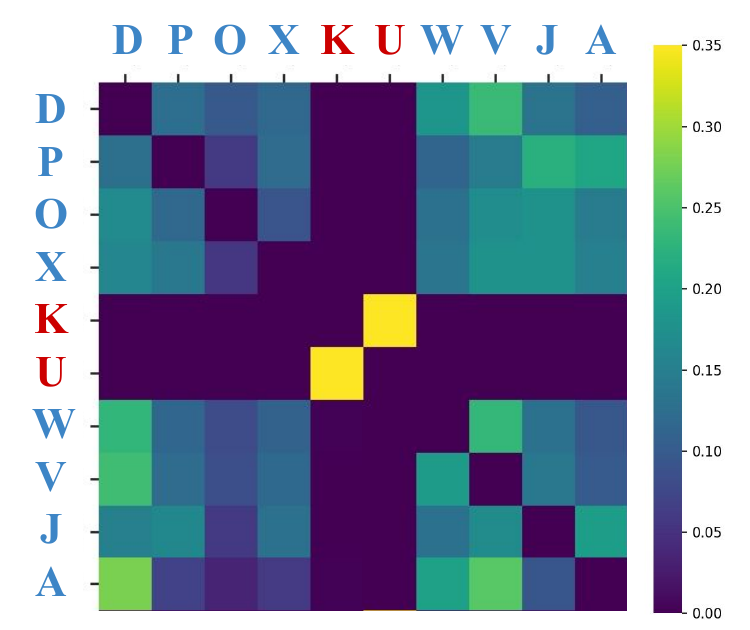}
        \caption{Attention map on the input prefix (all nodes except for the \textbf{target node}) demonstrating learned message passing.}
        \label{fig:attn_patterns}
    \end{subfigure}
    \caption{Qualitative analysis of GTLM's attention scores on a synthetic graph.}
    \label{fig:message_passing}
\end{figure}

To illustrate the internal mechanisms of GTLM, we fine-tune the 1-billion-parameter model on a synthetic task where it must predict whether two nodes belong to the same connected component (Figure~\ref{fig:clique_task}). By analyzing the resulting attention weights, we observe that many attention heads learn to implicitly simulate message passing by leveraging our structural attention biases (Figure~\ref{fig:attn_patterns}).

Note that the attention scores between nodes \textbf{K} and \textbf{U} and the remaining nodes are near zero, while the attention scores between same-clique nodes are higher. This experiment suggests that the GTLM architecture can natively process graphs akin to dedicated GNNs or GTs. This property explains the model's exceptionally strong performance on the GraphQA benchmark, as it tests the model's algorithmic reasoning abilities. Further details about this experiment are provided in Appendix~\ref{app:simulated-message-passing}.

\subsubsection*{Efficiency}

\textbf{Parameter Efficiency.} GTLM requires two types of parameters to be trained: (1) LoRA parameters, and (2) bias-related parameters. LoRA adapters are known to add only a few percent of the total base model parameters, specifically, around $2\%$ when training a 1-billion-parameter model with rank $32$. On the other hand, our bias-related parameter count is almost negligible in comparison to the total number of parameters in modern LLMs. In total, the 173,056 bias-related parameters---comprising 4,096 for SPD, 50,688 for RRWP, and 118,272 for the Magnetic Laplacian---account for only around $0.015\%$ of the 1-billion-parameter base LLM.

\textbf{Implementation.} Since the graph bias is an additive term on the attention logits, it renders the FlashAttention \cite{dao2022flashattention} backend unusable for GTLM. Therefore, we implement the graph-bias-compatible attention backend using a PyTorch FlexAttention \texttt{score\_mod} \citep{dong2025flex}. Unlike the naive eager approach, our implementation does not materialize the full $L \times L$ attention matrix $\bm{A}$, nor the full token-level bias matrix; it only instantiates the node-level bias matrix $[b(u,v)]_{u,v} \in \mathbb{R}^{N \times N}$ per head and layer. We use a sequence-to-node index to perform a single lookup of the appropriate bias value inside the kernel. Since each node contributes many tokens, $N \ll L$, and this removes the $O(L^2)$ memory term the bias introduced. On our TAG benchmarks, a forward+backward step costs around $2\times$ that of a standard LLM using Flash Attention on exactly identical token sequences with no graph machinery, at exactly the same peak memory. Compared with the eager implementation, the same step time is $2\mbox{-}10\times$ faster with substantially lower peak memory, removing the $O(L^2)$ bottleneck. Since the advantage only becomes meaningful at a scale beyond a few hundred tokens, we select the appropriate backend per experiment. More information and detailed scaling measurements are available in Appendix \ref{app:flex-attention}.

\subsubsection*{Future Research}

\textbf{Generalist Foundation Models.} Given our strong results across a wide variety of problem types related to graphs and text, alongside the theoretical guarantee that GTLM reduces to a standard LLM with text-only inputs, we believe GTLM to be a strong architecture for building a generalist foundation model for these combined modalities. This endeavor would require a large-scale corpus of various graph-related tasks for continual pretraining of the model, as well as an instruction-tuning set with graph-conditioned questions to facilitate zero-shot interactions. Ultimately, such a system could meaningfully accelerate progress in domains such as molecular discovery, logical reasoning, or enterprise knowledge retrieval, where data fundamentally is structured as a graph.

\section{Conclusion}
We introduced \textbf{GTLM}, a novel architecture that natively unifies graph and text modalities without relying on lossy node compression or disjointed multi-step pipelines. By mathematically aligning structural graph attention biases with the prefix-LM masking strategy, GTLM retains exact backward compatibility with its pretrained LLM foundation while achieving state-of-the-art performance on a variety of benchmarks with structured data. Beyond standard benchmark performance, we show that GTLM generalizes significantly better to longer out-of-distribution sequences than an LLM under a needle-in-a-graph multi-hop retrieval setting.

\bibliography{iclr2027_conference}
\bibliographystyle{iclr2027_conference}

\appendix

\section{Derivations of Structural Features}
\label{app:structural_derivations}

This appendix provides the formal mathematical definitions for the structural features utilized by our Relative Positional Encodings.

\subsection{Relative Random Walk Probabilities (RRWP)}
To capture complex structural dynamics, we utilize Relative Random Walk Probabilities \citep{ma2023graphinductivebiasestransformers}. For a given graph with an adjacency matrix $\mathbf{A}$ and a diagonal degree matrix $\mathbf{D}$, the transition matrix is defined as $\mathbf{M} := \mathbf{D}^{-1}\mathbf{A}$. The $K$-step walk probabilities between nodes $u$ and $v$ are formulated as:
\[ \mathbf{P}_{u,v} = [\mathbf{I}, \mathbf{M}, \mathbf{M}^2, \dots, \mathbf{M}^{K-1}]_{u,v} \in \mathbb{R}^K. \]
These features are subsequently processed via a learned MLP into dense scalar attention biases.

\subsection{The Magnetic Laplacian}
To capture directed topology, we utilize the Magnetic Laplacian \citep{magnet_citation_zhang}. We first define the symmetrized adjacency matrix as $\mathbf{A}_s(u, v) := \frac{1}{2}(\mathbf{A}(u, v) + \mathbf{A}(v, u))$. The corresponding degree matrix is defined as $\mathbf{D}_s(u, u) := \sum_{v \in \mathcal{V}} \mathbf{A}_s(u, v)$. 

Directional information is captured via a phase matrix $\mathbf{\Theta}^{(q)}$, defined as $\mathbf{\Theta}^{(q)}(u, v) := 2\pi q (\mathbf{A}(u, v) - \mathbf{A}(v, u))$ for $q \ge 0$. The complex Hermitian adjacency matrix is then given by $\mathbf{H}^{(q)} := \mathbf{A}_s \odot \exp(i\mathbf{\Theta}^{(q)})$, where $\odot$ denotes componentwise multiplication. From this, the normalized Magnetic Laplacian is defined as:
\[ \mathbf{L}_N^{(q)} := \mathbf{I} - (\mathbf{D}_s^{-1/2} \mathbf{A}_s \mathbf{D}_s^{-1/2}) \odot \exp(i\mathbf{\Theta}^{(q)}). \]

This matrix is Hermitian, positive-semidefinite, and thus diagonalized by an orthonormal basis of complex eigenvectors associated with real, nonnegative eigenvalues. Its spectral decomposition is given by $\mathbf{L}_N^{(q)} = \mathbf{V}\mathbf{\Lambda}\mathbf{V}^\dagger$. To resolve the unitary basis ambiguity in the complex eigenvectors $\mathbf{V}$, we follow an approach similar to \citet{huang2025what}, in order to construct a basis-invariant kernel:
\[ \mathbf{K} = \mathbf{V} \text{diag}(\phi(\lambda)) \mathbf{V}^\dagger \in \mathbb{C}^{N \times N \times d_{\text{Mag}}}, \]
where $\phi(\lambda)$ is a learned permutation-invariant transformation of the eigenvalues. This invariant kernel provides a mathematically robust foundation for extracting relative spectral distances between nodes.

\section{Theoretical Proofs and Empirical Verification of Architectural Properties}
\label{app:properties-proof}

In this section, we provide the formal mathematical proofs for the architectural guarantees introduced in Section~\ref{sec:theoretical-properties}, followed by their corresponding empirical and numerical validations.

\subsection{Formal Mathematical Proofs}

\textbf{Proof of Property 1: Prefix Node Permutation Equivariance}

\textit{Claim:} Let $\mathbf{X} = [\mathbf{x}_{v_1} \oplus \dots \oplus \mathbf{x}_{v_N}]$ be the serialized sequence of token embeddings for the $N$ prefix context nodes. GTLM is equivariant with respect to any permutation $\pi \in S_N$ applied to the order of these nodes, meaning $\text{GTLM}(\pi \cdot \mathbf{X}) = \pi \cdot \text{GTLM}(\mathbf{X})$, and strictly invariant with respect to the generated target sequence.

\textit{Proof:} In a standard Transformer layer, the updated representation for a token $i$ inside node $u$ is computed as a weighted sum of value vectors $V_j$ from tokens $j \in v$, weighted by the attention matrix $\text{softmax}(A_{i,j})$. Equivariance holds if and only if the set of attended tokens and their corresponding attention scores $A_{i,j}$ remain completely independent of the permutation $\pi$. We demonstrate this across the three relevant components of the attention mechanism:
\begin{enumerate}
    \item \textbf{Attention Masking:} GTLM applies a fully-visible, non-causal mask over the entire prefix. Therefore, the set of tokens that token $i$ attends to is exactly the entire prefix set $\bigcup_{k=1}^N \{t \in v_k\}$, which is invariant to the serialization order $\pi$.
    \item \textbf{Positional Encodings (RoPE):} In GTLM, the 1D positional index for RoPE is reset to $0$ at the start of every node. Consequently, the query $Q_i$ and key $K_j$ representations depend exclusively on their local token indices within their respective nodes $u$ and $v$, strictly decoupling them from the global flattened sequence index imposed by $\pi$.
    \item \textbf{Structural Biases:} The raw attention score is $A^{(l, h)}_{i,j} = \frac{Q_i \cdot K_j^\top}{\sqrt{d_{\text{head}}}} + b^{(l,h)}(u,v)$. The bias term $b^{(l,h)}(u,v) = b_{\text{SPD}} + b_{\text{RRWP}} + b_{\text{Mag}}$ is computed using static adjacency matrices, transition probabilities, and Laplacian spectra derived from the immutable graph topology $\mathcal{G}$. These graph-level properties are inherently permutation invariant.
\end{enumerate}
Because the mask, the representations $Q$ and $K$, and the structural biases $b(u,v)$ are all mathematically invariant to $\pi$, the computed attention scores $A_{i,j}$ for any given pair of tokens remain identical regardless of the node ordering. Thus, the resulting token representations are identical in value and simply undergo the same permutation $\pi$ as the input, satisfying exact permutation equivariance for the prefix context. $\blacksquare$

\vspace{1em}
\textbf{Proof of Property 2: Pretraining Backward Compatibility}

\textit{Claim:} For any graph $\mathcal{G}$ consisting of a single node (representing a standard text sequence), the GTLM forward pass is mathematically identical to that of the unmodified base LLM.

\textit{Proof:} Consider a single-node graph $\mathcal{G}=(\{u\}, \emptyset)$. For any two tokens $i$ and $j$, we have $u=v$, and the attention score between those two tokens within the node $u$ is defined as:
\[
A^{(l, h)}_{i,j} = \frac{(Q^{(l, h)}_i) \cdot (K^{(l, h)}_j)^\top}{\sqrt{d_{\text{head}}}} + b_{\text{SPD}}^{(l,h)}(u,u) + b_{\text{RRWP}}^{(l,h)}(u,u) + b_{\text{Mag}}^{(l,h)}(u,u).
\]
By the architectural definitions established in Section \ref{sec:modified-attention}, each structural bias natively evaluates to zero for intra-node interactions:
\begin{enumerate}
    \item $b_{\text{SPD}}^{(l,h)}(u,u) = 0$, as the shortest path distance from a node to itself is exactly $0$, which is explicitly mapped to a zero-bias vector.
    \item $b_{\text{RRWP}}^{(l,h)}(u,u) = 0$, enforced by the piecewise formulation where $u=v$.
    \item $b_{\text{Mag}}^{(l,h)}(u,u) = 0$, enforced by the piecewise formulation where $u=v$.
\end{enumerate}
Summing these terms completely cancels the structural biases, reducing the computation to $A^{(l, h)}_{i,j} = \frac{Q_i \cdot K_j^\top}{\sqrt{d_{\text{head}}}}$. Furthermore, because the sequence consists of only one node, the per-node RoPE reset behaves identically to standard RoPE over a single continuous sequence. Causal masking constraints apply to the target sequence exactly as they do in the base model. Thus, all matrix operations in the attention mechanism exactly mirror those of standard scaled dot-product attention, guaranteeing zero degradation of pretrained capabilities. $\blacksquare$

\subsection{Empirical Verification}

To complement the theoretical proofs provided above, we numerically verify both architectural guarantees. Because these properties are fundamental to the model's forward pass and operate independently of the learned weights, we evaluate them using the base \texttt{Llama-3.2-1B} model with randomly initialized bias-related parameters.

\textbf{Backward Compatibility.} We construct a trivial graph consisting of a single \textbf{target node} with no edges, assigned a randomly generated textual attribute. We perform a forward pass on this input using both our GTLM architecture and the unmodified base Llama model. We monitor the maximum absolute difference between the output logits of the two models. Across five independent trials, we found this difference to be $2.1 \times 10^{-5} \pm 8.3\times 10^{-6}$. This delta is entirely within the expected floating-point (FP32) accumulation error, numerically confirming exact backward compatibility.

\textbf{Permutation Equivariance.} We construct a multi-node test graph with a fixed topology. We then generate two flattened input sequences by randomly permuting the serialization order of the prefix nodes, while strictly preserving their textual attributes and the underlying graph structure. Both permuted sequences are processed by GTLM, and the outputs are inversely permuted to align with the canonical node order. We measure the maximum absolute difference in the final logits between the two permuted sequences. The result we get across five independent trials is $2.77 \times 10^{-5} \pm 2.87\times 10^{-6}$. These results confirm that our architecture is strictly equivariant to the arbitrary ordering of nodes in the flattened sequence.

Code to reproduce these numerical verification checks is available in our codebase alongside the main training framework. Note that we report the numbers as mean $\pm$ sample standard deviation.

\section{General GTLM Training Setup}
\label{app:experimental-details}

\textbf{Attention Bias Hyperparameters.} Unless stated otherwise, all GTLM experiments employ a fixed configuration for the structural attention biases. Specifically, we fix the maximum distance for Shortest Path Distance (SPD), the maximum steps for Relative Random Walk Probabilities (RRWP), and the $q$ value and dimensionality for Magnetic Laplacian edge-level embeddings. The parameter values are shown in Table~\ref{tab:bias_parameter_vals}.

\begin{table}[ht]
    \centering
    \small
    \caption{Graph-aware attention bias parameter values.}
    \label{tab:bias_parameter_vals}
    \vspace{0.5em}
    \begin{tabular}{l r}
        \toprule
        \textbf{Parameter} & \textbf{Value} \\
        \midrule
        SPD max dist.       & 8 \\
        max RRWP steps      & 16 \\
        Magnetic Lap. $q$   & 0.25 \\
        Magnetic Lap. dim.  & 32 \\
        \bottomrule
    \end{tabular}
\end{table}

\textbf{Learning rates.} We utilize a differential learning rate strategy, assigning a higher rate to the structural attention bias parameters than to the LoRA adapters. This approach is motivated by two factors: (1) the bias parameters are initialized randomly and must learn to map graph topology to the attention space, whereas the LLM and adapters start from highly refined, pretrained weights, and (2) despite their low parameter count, these biases are the primary mechanism for graph-awareness and require a stronger gradient signal to converge. This ensures that the essential structural features are learned effectively without being overshadowed by the larger, pretrained components during optimization.

\section{GTLM with ALiBi and dual RoPE}
\label{app:other-pos-encodings}

In order to demonstrate the generality of the GTLM architecture, specifically the per-node position index resetting idea, we show that it is possible to implement with \textbf{ALiBi} and \textbf{dual RoPE}. We use BLOOM \citep{scao2022bloom} as base model for the former, and Gemma-3-1B \citep{gemma3report2025} for the latter.

\textbf{ALiBi.} ALiBi adds a fixed distance penalty per head to the attention logits before softmax. GTLM's structural mask and graph bias are added in the same slot, so BLOOM's attention requires no changes. The only adaptation is to calculate ALiBi distances from GTLM's node-position identifiers instead of the raw token order. This prevents the node order from acting as a spurious prior, and the single-node graph still accurately reproduces standard BLOOM.

\textbf{Dual RoPE.} GTLM only changes the position IDs provided to RoPE, not the rotation, so it does not depend on the underlying frequency of RoPE. The local (10k) and global (1M) Gemma-3 layers rotate the same position identifiers per node. Each layer therefore encodes the order of tokens within a node, and the graph bias, added identically in each layer, carries the structure between nodes. The 512-token sliding window is discarded because it follows the serialization order, not the graph. At GraphQA sequence lengths (maximum 126 tokens), this change has no effect.

We provide the results for both and compare them to the results obtained with Llama-3.2-1B as the base model (Table \ref{tab:other-pos-encodings}).

\begin{table}[ht]
    \caption{GraphQA per-task accuracy ($\%$) across GTLM base models with different positional encodings (\textbf{best}, \underline{second-best}).}
    \label{tab:other-pos-encodings}
    \vspace{0.5em}
    \centering
    \renewcommand{\arraystretch}{1.1}
    
    \setlength{\tabcolsep}{8pt} 
    
    \resizebox{\textwidth}{!}{
    \begin{tabular}{@{}l *{4}{c} *{2}{c} *{3}{c}@{\hspace{1.8em}}c@{}}
        \toprule
        & \multicolumn{4}{c}{Graph Tasks} & \multicolumn{2}{c}{Node Tasks} & \multicolumn{3}{c@{\hspace{1.8em}}}{Edge Tasks} & \\
        \cmidrule(lr){2-5} \cmidrule(lr){6-7} \cmidrule(lr{1.8em}){8-10}
        & Node & Edge & Cycle & Triangle & Node & Conn. & Reach- & Edge & Shortest & \\
        GTLM Base Model & Count & Count & Check & Count & Degree & Nodes & ability & Exist. & Path & Avg. \\
        \midrule
        Llama-3.2-1B {\scriptsize(RoPE)}      & \underline{99.7}\sstd{0.5} & \textbf{57.2}\sstd{2.7}    & \textbf{97.7}\sstd{0.8}    & \textbf{31.5}\sstd{0.8}    & \textbf{99.9}\sstd{0.1} & \underline{94.3}\sstd{0.1} & \textbf{98.4}\sstd{0.0}    & \textbf{99.9}\sstd{0.1}    & \textbf{92.7}\sstd{1.7}    & \textbf{85.7} \\
        bloom-1b1 {\scriptsize(ALiBi)}        & \textbf{100}\sstd{0.0}     & \underline{41.5}\sstd{2.8} & \underline{97.5}\sstd{1.3} & \underline{31.2}\sstd{0.8} & 94.4\sstd{3.1}          & \textbf{98.5}\sstd{0.5}    & \textbf{98.4}\sstd{0.2}    & \underline{99.7}\sstd{0.1} & \textbf{92.7}\sstd{0.8}    & \underline{83.8} \\
        gemma-3-1b-pt {\scriptsize(dual RoPE)} & \textbf{100}\sstd{0.0}     & 39.7\sstd{2.0}             & 97.3\sstd{0.5}             & 25.8\sstd{1.0}             & \underline{95.6}\sstd{1.4} & 85.5\sstd{1.1}          & \underline{98.3}\sstd{0.3} & \textbf{99.9}\sstd{0.1}    & \underline{92.2}\sstd{0.7} & 81.6 \\
        \bottomrule
    \end{tabular}
    }
\end{table}

\section{GraphQA}
\label{app:graphqa}

The \textbf{incidence graph} is a transformation of a directed graph $\mathcal{G}=(\mathcal{V},\mathcal{E})$ into a bipartite directed graph $\overline{\mathcal{G}}=(\overline{\mathcal{V}}, \overline{\mathcal{E}})$ as follows:
\[ \overline{\mathcal{V}} = \mathcal{V} \; \cup \; \mathcal{E} \]
\[ \overline{\mathcal{E}} = \{ (u, e), (e, v) \mid e = (u, v) \in \mathcal{E} \} \]

In this bipartite formulation, every original directed edge $e=(u,v)$ is effectively represented as a node in $\overline{\mathcal{V}}$ and is replaced by a path of length two: $u \to e \to v$. This transformation ensures that both vertices and edges from the original graph are treated uniformly as nodes in the incidence graph. Consequently, any edge attributes or topological properties can be seamlessly processed by GTLM's structural attention biases alongside the node data.

\textbf{Graph Construction}. We use the graphs from the publicly available GraphQA dataset and set the textual attributes of each node to its own unique ID. When using incidence graphs, the original edges are given the text attributes with format "($i$, $j$)", where $i$ and $j$ are the nodes the given edge connects.

\textbf{Training Hyperparameters} used to train GTLM on the \textit{standard} and \textit{incidence} graphs are the same and shown in Table~\ref{tab:graphqa-hyperparams}.

\begin{table}[h]
\centering
\small
\caption{Hyperparameter values used to train GTLM on the GraphQA benchmark.}
\label{tab:graphqa-hyperparams}
\vspace{0.5em}
\begin{tabular}{@{}lc@{}}
\toprule
\textbf{Parameter} & \textbf{Value} \\ \midrule
Epochs & $20$ \\
Learning Rate ($\eta$) & $3 \times 10^{-5}$ \\
Bias Learning Rate ($\eta_{\text{bias}}$) & $5 \times 10^{-3}$ \\
LoRA Rank ($r$) & $16$ \\
LoRA Alpha ($\alpha$) & $32$ \\ \bottomrule
\end{tabular}
\end{table}

\section{TAG Benchmarks}
\label{app:tag-benchmarks}

\textbf{Dataset splits.} The dataset splits for these four benchmarks were taken from \citet{zhang2026graphtokenizinglargelanguagemodels} to ensure a fair comparison to the RGLM models and their baselines.

\textbf{Neighborhood sampling.} To construct the inputs for GTLM, we construct ego subgraphs for each node with up to $N$ ($30$ or $60$) 2-hop neighbors. In cases when there were more than $N$ 2-hop neighbors, they were chosen based on closeness, while equidistant nodes were sampled randomly.

\textbf{Text attributes.} For the three benchmarks with paper citation graphs we employ the following strategy: the target paper contains the full title and abstract, while all others contain only the title of the paper. On the other hand, after some experimenting, we found that simply truncating the post contents in the \textit{Reddit} dataset to around $25$ tokens per node yielded strong results while maintaining manageable context lengths.

\textbf{Question + Answer injection.} In order to train the model for autoregressive generation solely on the \texttt{label} textual sequence, we append the following string to the end of the target node: \\
\texttt{\textbackslash{}n\textbackslash{}n\{task\_question\}\textbackslash{}n A: \{label\}}

\textbf{Training Hyperparameters.} We conducted a search over multiple parameters with all options shown and the chosen setup highlighted, as seen in Table~\ref{tab:tag-hyperparams}. For the neighborhood sizes we chose between $30$ and $60$ based on the context length such that the average total sequence length did not exceed $1000$ tokens, which was a heuristic choice to balance training speeds and performance.

\begin{table}[ht]
\centering
\small
\caption{\textbf{Hyperparameter Configurations.} Final values and search spaces for each dataset. For learning rates and LoRA rank, the search set is provided with the selected optimal value \textbf{bolded}.}
\label{tab:tag-hyperparams}
\vspace{0.5em}
\begin{tabular}{@{}l cccc@{}}
\toprule
\textbf{Parameter} & \textbf{Cora} & \textbf{PubMed} & \textbf{ogbn-arxiv} & \textbf{Reddit} \\ \midrule
Neighbors ($k$) & 60 & 30 & 60 & 30 \\
Epochs & 20 & 4 & 1 & 15 \\
Learning Rate ($\eta$) & \{6e-5, 2e-4, \textbf{3e-4}\} & \{6e-5, \textbf{2e-4}, 3e-4\} & \{6e-5, \textbf{2e-4}, 3e-4\} & \{\textbf{6e-5}, 2e-4, 3e-4\} \\
Bias LR ($\eta_{\text{bias}}$) & \{0.01, \textbf{0.04}\} & \{0.01, \textbf{0.04}\} & \{0.01, \textbf{0.04}\} & \{\textbf{0.01}, 0.04\} \\
LoRA Rank ($r$) & \{32, \textbf{64}\} & \{\textbf{32}, 64\} & \{32, \textbf{64}\} & \{\textbf{32}, 64\} \\
LoRA Alpha ($\alpha$) & $2r$ & $2r$ & $2r$ & $2r$ \\ \bottomrule
\end{tabular}
\end{table}

\section{GraphRAG}
\label{app:graphrag}

As input to GTLM we use the SR subgraphs \citep{sr2022}, which we downloaded from GNN-RAG's public Google Drive folder\footnote{\texttt{data.zip} and \texttt{entities\_names.json} files from: \url{https://drive.google.com/drive/folders/1ifgVHQDnvFEunP9hmVYT07Y3rvcpIfQp}}.
We convert each subgraph into a directed Levi graph, in which every triple $(h, r, t)$ becomes a path $h \rightarrow r \rightarrow t$. Entity nodes are verbalized by their Freebase names from \texttt{entities\_names.json}, which we extend with Freebase name aliases, and relation nodes by the last segment of the relation identifier (e.g.\ \texttt{people.person.place\_of\_birth} becomes ``place of birth''); graphs are capped at 512 nodes for WebQSP and 1024 for CWQ.
The question is placed in its own node in the prefix with no edges to the rest of the graph, so the structural encodings treat it as a separate component while every graph token can still attend to it, and the answer is generated from a separate prompt (target) node connected to the topic entities.
We use the standard train, dev and test splits of both benchmarks: training keeps only questions whose answer appears in the retrieved subgraph, while dev and test keep every question and score unanswerable ones as 0, so our evaluation denominators match those of RoG and GNN-RAG.

Table~\ref{tab:kgqa-hyperparams} lists the hyperparameters required to reproduce our results with the 1B (Llama-3.2-1B) and 8B (Llama-3.1-8B) base models on WebQSP and CWQ. For hyperparameters that were tuned, the table gives every value we tried and marks the one we selected.

\begin{table}[ht]
    \centering
    \small
    \caption{\textbf{KGQA Hyperparameter Configurations.} Final values and search spaces for WebQSP and CWQ across 1B and 8B backbones. Where a search set is provided, the selected optimal value is \textbf{bolded}.}
    \label{tab:kgqa-hyperparams}
    \vspace{0.5em}
    \setlength{\tabcolsep}{10pt}
    \begin{tabular}{@{}l cc cc@{}}
        \toprule
         & \multicolumn{2}{c}{\textbf{Llama-3.2-1B}} & \multicolumn{2}{c}{\textbf{Llama-3.1-8B}} \\
        \cmidrule(lr){2-3} \cmidrule(lr){4-5}
        Parameter & \textbf{WebQSP} & \textbf{CWQ} & \textbf{WebQSP} & \textbf{CWQ} \\
        \midrule
        Max Nodes & 512 & 1024 & 512 & 1024 \\
        Max Answers ($n_{\max}$) & 50 & 50 & 50 & 50 \\
        Epochs & 15 & 8 & 15 & 8 \\
        Learning Rate ($\eta$) & \{5e-5, \textbf{1e-4}\} & \{5e-5, \textbf{1e-4}\} & \{\textbf{5e-5}, 1e-4\} & \{\textbf{5e-5}, 1e-4\} \\
        Bias LR ($\eta_{\text{bias}}$) & \{\textbf{5e-3}, 1e-2\} & 5e-3 & 5e-3 & 5e-3 \\
        LoRA Rank ($r$) & \{16, \textbf{64}\} & 64 & \{16, \textbf{64}\} & 64 \\
        LoRA Dropout & \{\textbf{0.15}, 0.25\} & 0.15 & 0.15 & 0.15 \\
        \bottomrule
    \end{tabular}
\end{table}

Table~\ref{tab:kgqa-ceilings} gives the upper bound on the response coverage for each input condition on the test split, i.e. the best result that any reader can achieve: the Hits@1 ceiling is the fraction of questions with at least one gold answer in the input graph, and the F1 ceiling is the macro F1 of a reader that outputs exactly the gold answers present. Our node cap costs at most 0.8 points. On WebQSP, the 2-hop pool that RoG and GNN-RAG retrieve from has a 5.2-point higher F1 ceiling than our SR inputs, while on CWQ the two ceilings are almost equal.

\begin{table}[ht]
    \centering
    \small
    \caption{\textbf{KGQA Answer-Coverage Ceilings} on the test split (\%).}
    \label{tab:kgqa-ceilings}
    \vspace{0.5em}
    \begin{tabular}{@{}l cc cc@{}}
        \toprule
         & \multicolumn{2}{c}{\textbf{WebQSP}} & \multicolumn{2}{c}{\textbf{CWQ}} \\
        \cmidrule(lr){2-3} \cmidrule(lr){4-5}
        Input condition & Hits@1 & F1 & Hits@1 & F1 \\
        \midrule
        RoG / GNN-RAG 2-hop pool & 95.6 & 94.3 & 80.7 & 79.5 \\
        SR, uncapped & 91.1 & 89.6 & 80.7 & 79.8 \\
        SR, as used (512 / 1024 nodes) & 90.9 & 89.1 & 79.9 & 79.0 \\
        \bottomrule
    \end{tabular}
\end{table}

\section{Molecule Benchmarks}
\label{app:molecule-benchmarks}

For property prediction, we use the BACE, BBBP, and HIV classification sets from MoleculeNet \citep{wu2018moleculenet}, split using DeepChem's deterministic Bemis--Murcko scaffold splitter (80/10/10), the split used by the baselines we compare against. For captioning, we use ChEBI-20 \citep{edwards2021text2mol} with its own splits for training, validation, and testing\footnote{\texttt{ChEBI-20\_data} directory \url{https://github.com/blender-nlp/MolT5}}.
We parse each SMILES string using RDKit into a heavy-atom graph and convert it into a Levi graph, where atoms and bonds are nodes, and each atom and bond incidence is an edge in both directions. Atom nodes are verbalized by the features a standard GNN featuriser receives, namely element, aromaticity, ring membership, heavy-atom degree, hydrogen count, formal charge, and chiral tag (e.g., ``carbon aromatic ring deg3 H1''), and bond nodes by their bond order and ring membership (e.g., ``double bond in ring''), so the model never sees a SMILES string.
Similar to the GraphRAG experiments, the question is placed in its own node in the graph's prefix with no edges to the rest of the graph, and the answer is generated from a separate prompt (target) node, which here is connected to every atom node because the questions concern the molecule as a whole.
Property prediction is posed as a yes/no question (e.g.\ ``Does this molecule penetrate the blood-brain barrier?''), and we score ROC-AUC from the margin between the ``Yes'' and ``No'' logits at the answer position; captioning answers ``Describe this molecule.'' with the reference caption.

Tables~\ref{tab:molprop-hyperparams} and~\ref{tab:chebi-hyperparams} list the hyperparameters required to reproduce our results with the 1B (Llama-3.2-1B) and 8B (Llama-3.1-8B) base models. 
All runs use a one-epoch linear warm-up, followed by a cosine decay to one-tenth of the peak learning rate. For property prediction, we evaluate on the validation split at a fixed step interval and report the test result of the checkpoint with the best validation ROC-AUC; for captioning, we report the final checkpoint. For the hyperparameters that are tuned, the tables give each value we tried and indicate the one we chose.

\begin{table}[ht]
    \centering
    \small
    \caption{\textbf{Property Prediction Hyperparameter Configurations.} Final values and search spaces for BACE, BBBP and HIV across 1B and 8B backbones. Where a search set is provided, the selected value is \textbf{bolded}.}
    \label{tab:molprop-hyperparams}
    \vspace{0.5em}
    \setlength{\tabcolsep}{4pt}
    \resizebox{\textwidth}{!}{%
    \begin{tabular}{@{}l ccc ccc@{}}
        \toprule
         & \multicolumn{3}{c}{\textbf{Llama-3.2-1B}} & \multicolumn{3}{c}{\textbf{Llama-3.1-8B}} \\
        \cmidrule(lr){2-4} \cmidrule(lr){5-7}
        Parameter & \textbf{BACE} & \textbf{BBBP} & \textbf{HIV} & \textbf{BACE} & \textbf{BBBP} & \textbf{HIV} \\
        \midrule
        Epochs & 40 & 40 & 10 & 40 & 40 & 10 \\
        Global Batch Size & 32 & 32 & 32 & 32 & 32 & 32 \\
        Learning Rate ($\eta$) & \{3e-5, \textbf{3e-4}\} & \{3e-5, \textbf{3e-4}\} & \{3e-5, \textbf{1e-4}, 3e-4\} & \{1.5e-4, \textbf{3e-4}\} & \{\textbf{1.5e-4}, 3e-4\} & \{\textbf{5e-5}, 1e-4\} \\
        Bias LR ($\eta_{\text{bias}}$) & 1e-2 & 1e-2 & 1e-2 & 1e-2 & 1e-2 & 1e-2 \\
        LoRA Rank ($r$) & \{\textbf{16}, 32\} & \{\textbf{16}, 32\} & \{\textbf{16}, 32\} & 16 & 16 & 16 \\
        LoRA Dropout & 0.05 & 0.05 & 0.05 & 0.05 & 0.05 & 0.05 \\
        Max SPD Distance & \{\textbf{32}, 64\} & \{\textbf{32}, 64\} & \{\textbf{32}, 64\} & 32 & 32 & 32 \\
        \bottomrule
    \end{tabular}%
    }
\end{table}

\begin{table}[ht]
    \centering
    \small
    \caption{\textbf{ChEBI-20 Hyperparameter Configurations.} Final values and search spaces for molecule captioning across 1B and 8B backbones. Where a search set is provided, the selected value is \textbf{bolded}.}
    \label{tab:chebi-hyperparams}
    \vspace{0.5em}
    \setlength{\tabcolsep}{10pt}
    \begin{tabular}{@{}l cc@{}}
        \toprule
        Parameter & \textbf{Llama-3.2-1B} & \textbf{Llama-3.1-8B} \\
        \midrule
        Max Heavy Atoms & 128 & 128 \\
        Epochs & 12 & 12 \\
        Batch Size & 64 & 64 \\
        Learning Rate ($\eta$) & \{5e-5, 2e-4, 3e-4, 5e-4, \textbf{1e-3}\} & \{1e-4, 3e-4, \textbf{5e-4}\} \\
        Bias LR ($\eta_{\text{bias}}$) & \{\textbf{5e-3}, 2e-2\} & 5e-3 \\
        LoRA Rank ($r$) & 16 & 16 \\
        LoRA Dropout & 0.05 & 0.05 \\
        Max SPD & 32 & 32 \\
        \bottomrule
    \end{tabular}
\end{table}

\section{Needle-in-a-Graph Experiment}
\label{app:context-experiment}

\paragraph{Task.}
Each graph has $N$ nodes: a question node, $N{-}2$ content nodes, and a prompt (target) node where the answer is generated.
Each content node $v$ holds exactly $T$ tokens of WikiText-103 \citep{merity2017pointer} filler text with one record sentence inserted at a random offset:
\begin{center}
\small\texttt{The access code for $\mathrm{id}(v)$ is $\mathrm{code}(v)$. Continue at $\mathrm{id}(\mathrm{succ}(v))$. Decoy reference: $\mathrm{id}(\mathrm{dec}(v))$.}
\end{center}
The question asks ``\emph{Starting at $\mathrm{id}(c_0)$, follow $k$ `Continue at' references, ignoring any decoy references. What is the access code there?}'', and the target is $\mathrm{code}(c_k)$ with $c_{i+1} = \mathrm{succ}(c_i)$.
Both pointers (the correct and decoy) become untyped directed edges $v \rightarrow \mathrm{succ}(v)$ and $v \rightarrow \mathrm{dec}(v)$, so about $2^k$ nodes lie at distance $k$ from $c_0$ and the path can only be resolved by reading the node texts. An oracle using only shortest-path distances scores $15.6\%$ at $N{=}32$, $k{=}3$, where GTLM scores $99\%$, proving GTLM could not achieve those results with just an $O(1)$ retrieval by decoding structural features. Furthermore, we ensure that the full graph topology is fully derivable from the text the LLM receives, making the comparison fair.
The question node is connected to every content node, and the prompt node has no edges.

\paragraph{Data.}
Identifiers (e.g.\ \texttt{NODE-0728}) are sampled without replacement from a pool of $4{,}096$. Codes follow the pattern \texttt{LLDLL} (e.g.\ \texttt{AB1CD}) and tokenize to exactly three tokens.
To make the cells comparable, each test graph is generated once and reused across the grid. A larger $N$ adds nodes around the same path, a larger $k$ extends the same path from the same starting node, and a different $T$ only changes the amount of padding around each record. Node order is shuffled.
The $16{,}000$ training graphs sample $k \in \{1,2,3,4\}$, then $N$, then $T$ uniformly among the cells whose input fits into $16{,}384$ tokens. The cells $(N,T) \in \{(64,512), (128,256), (128,512)\}$ ($32$k--$65$k tokens) are never seen in training (* in Table~\ref{tab:nt-scaling}).
Each $(N, T, k)$ cell is tested on $200$ graphs.

\paragraph{Models.}
Both models fine-tune Llama-3.2-1B with LoRA under the same budget (Table~\ref{tab:context-hyperparams}).
The LLM receives the node texts in random order, each preceded by \texttt{\#\# Article $i$}, followed by the question and \texttt{Answer:}.
We report exact match on the three code tokens, taking the greedy prediction at each answer position given the prefix, for the seed with the median validation accuracy out of three. Over all three seeds, the LLM averages $81.1\%$ ($k{=}2$) and $21.1\%$ ($k{=}4$) across the cells of Table~\ref{tab:nt-scaling}, against $89.9\%$ and $17.9\%$ for the median seed.

\begin{table}[ht]
    \centering
    \small
    \caption{\textbf{Needle-in-a-Graph Hyperparameters.} Both models use Llama-3.2-1B, a $10\%$ linear warm-up and cosine decay to one-tenth of the peak learning rate. We validate every $500$ steps and keep the best checkpoint.}
    \label{tab:context-hyperparams}
    \vspace{0.5em}
    \setlength{\tabcolsep}{10pt}
    \begin{tabular}{@{}l cc@{}}
        \toprule
        Parameter & \textbf{GTLM} & \textbf{LLM} \\
        \midrule
        Training Graphs / Epochs & 16,000 / 2 & 16,000 / 2 \\
        Max Training Length (tokens) & 16,384 & 16,384 \\
        Global Batch Size & 8 & 8 \\
        Learning Rate ($\eta$) & 1e-4 & 1e-4 \\
        Bias LR ($\eta_{\text{bias}}$) & 5e-3 & -- \\
        Weight Decay & 0.1 & 0.1 \\
        LoRA Rank / $\alpha$ / Dropout & 64 / 128 / 0.15 & 64 / 128 / 0.15 \\
        \bottomrule
    \end{tabular}
\end{table}

\paragraph{Node-order sensitivity.}
We re-evaluate the checkpoints of Table~\ref{tab:nt-scaling} on the same test graphs, changing only the node order.
The $k{+}1$ path nodes form a grouped block, in path order (fwd) or reversed (rev), inserted at one of five relative positions among the other nodes, whose relative order is unchanged. Together with the original random order, this gives $11$ orderings per graph.
For GTLM, reordering permutes the node indices together with the structural encodings and edges.

We use all $200$ test graphs, all three LLM seeds, and the reported GTLM seed. Because every graph is scored under every ordering, comparisons are paired. We report $95\%$ CIs from a bootstrap that resamples graphs (keeping a graph's seeds together), and $p$-values from exact McNemar tests on the graphs whose correctness changes between two orderings.

\begin{table}[ht]
    \centering
    \small
    \caption{\textbf{Node-order sensitivity.} Accuracy (\%) on the same $200$ test graphs per cell when only the node order changes. The path block is in path order (fwd) or reversed (rev); \emph{End} is adjacent to the question, and \emph{Random} is the order of Table~\ref{tab:nt-scaling}. LLM values are means over three seeds. GTLM scores identically under all $11$ orderings. * marks cells outside the training distribution.}
    \label{tab:node-order}
    \vspace{0.5em}
    \setlength{\tabcolsep}{5pt}
    \begin{tabular}{@{}cc c ccccc c c@{}}
        \toprule
         & & & \multicolumn{6}{c}{\textbf{LLM}} & \textbf{GTLM} \\
        \cmidrule(lr){4-9} \cmidrule(l){10-10}
        $k$ & $(N, T)$ & Block & Start & 0.25 & Middle & 0.75 & End & Random & All orders \\
        \midrule
        \multirow{6}{*}{2}
          & \multirow{2}{*}{$(32, 512)$}    & fwd & 91.8 & \textbf{94.0} & 90.2 & 83.7 & 80.8 & \multirow{2}{*}{84.2} & \multirow{2}{*}{100} \\
          &                                 & rev & 75.7 & 73.7 & 69.5 & 64.3 & 66.5 & & \\
          & \multirow{2}{*}{$(128, 128)$}   & fwd & \textbf{94.0} & 88.8 & 80.0 & 70.7 & 63.5 & \multirow{2}{*}{74.2} & \multirow{2}{*}{100} \\
          &                                 & rev & 87.3 & 52.8 & 45.0 & 37.3 & 40.8 & & \\
          & \multirow{2}{*}{$(128, 512)$*}  & fwd & \textbf{81.3} & 77.3 & 68.3 & 65.0 & 62.3 & \multirow{2}{*}{68.8} & \multirow{2}{*}{100} \\
          &                                 & rev & 76.8 & 32.2 & 29.3 & 38.3 & 45.0 & & \\
        \midrule
        \multirow{6}{*}{4}
          & \multirow{2}{*}{$(32, 512)$}    & fwd & \textbf{25.0} & 20.8 & 22.3 & 18.5 & 17.7 & \multirow{2}{*}{23.8} & \multirow{2}{*}{94.0} \\
          &                                 & rev & 5.8 & 5.8 & 5.7 & 7.0 & 10.2 & & \\
          & \multirow{2}{*}{$(128, 128)$}   & fwd & 10.5 & \textbf{11.5} & 8.0 & 5.2 & 5.7 & \multirow{2}{*}{12.0} & \multirow{2}{*}{96.5} \\
          &                                 & rev & 5.3 & 1.3 & 0.7 & 1.3 & 1.3 & & \\
          & \multirow{2}{*}{$(128, 512)$*}  & fwd & 4.7 & 6.0 & 5.7 & 6.0 & \textbf{8.7} & \multirow{2}{*}{7.3} & \multirow{2}{*}{97.5--98.0} \\
          &                                 & rev & 1.8 & 0.5 & 0.8 & 1.8 & 4.2 & & \\
        \bottomrule
    \end{tabular}
\end{table}

With the checkpoint and inputs fixed, node order alone changes the LLM's accuracy by up to $56.7$ percentage points (pp; Table~\ref{tab:node-order}, $k{=}2$, $(128,128)$).
Reversing the path costs between $4$ and $27$ pp, averaged over positions. Accuracy is generally highest with the path near the start of the context and lowest with it next to the question (up to $30.5$ pp lower), so the effect is not U-shaped as it would be in the \emph{lost in the middle} phenomenon.
These differences are significant in every cell: the $95\%$ CI of the reversal cost excludes zero everywhere, and start vs.\ end has $p < 0.004$ everywhere ($p < 10^{-7}$ in all $k{=}2$ cells).

At $k{=}2$, placing the path first recovers much of the loss from larger graphs: $(128,128)$ reaches $94.0\%$, above the LLM's $88.2\%$ average on the $N{=}16$ cells. At $k{=}4$, no ordering exceeds $25\%$.
GTLM's prediction is unchanged in $11{,}997$ of $12{,}000$ (graph, ordering) pairs, and its mean gold log-probability varies by at most $0.01$ nats, against $0.5$--$3.6$ for the LLM. This is the invariance of Property~1, up to floating-point summation order.

\section{Scaling GTLM to Larger Base Models}
\label{app:scaling-base}

In the process of optimizing GTLM's performance on the various benchmarks mentioned in this paper, we examined the dynamics of scaling the base model beyond 1B parameters. As seen in Table \ref{tab:scaling-results}, we find that GTLM benefits the most from scaling in cases when the task involves open-ended generation (tasks (2) and (4)), while it did not contribute in a statistically significant way with the classification tasks (1) and (3).

\begin{table}[ht]
\centering
\caption{GTLM scaling performance across Llama-3.2-1B, Llama-3.2-3B, and Llama-3.1-8B backbones. $\Delta_{\text{rel}}$ denotes the relative percentage improvement from 1B to 8B.}
\label{tab:scaling-results}
\small
\setlength{\tabcolsep}{14pt}
\begin{tabular}{@{}l@{\hspace{4em}} c@{\hspace{4.2em}} c@{\hspace{4.2em}} c@{\hspace{4.2em}} r@{}}
\toprule
Task / Metric & 1B & 3B & 8B & $\Delta_{\text{rel}}$ \\
\midrule
\multicolumn{5}{@{}l}{\textit{(1) TAG Node Classification}} \\
Cora (Acc)        & 88.99\sstd{1.12}  & 89.17\sstd{0.70} & 89.25\sstd{0.36}  & $+0.3\%$ \\
Cora (F1)         & 91.31\sstd{0.90}  & 91.58\sstd{0.97} & 91.60\sstd{0.53}  & $+0.3\%$ \\
Pubmed (Acc)      & 94.57\sstd{0.11}  & 94.52\sstd{0.28} & 94.67\sstd{0.08}  & $+0.1\%$ \\
Pubmed (F1)       & 94.76\sstd{0.10}  & 94.72\sstd{0.18} & 94.93\sstd{0.11}  & $+0.2\%$ \\
\midrule
\multicolumn{5}{@{}l}{\textit{(2) Knowledge Graph QA}} \\
WebQSP (Hit)        & 83.25\sstd{1.00}  & --               & 85.9             & $+3.2\%$ \\
WebQSP (H@1)        & 78.03\sstd{1.15}  & --               & 82.5             & $+5.7\%$ \\
WebQSP (F1)         & 73.51\sstd{0.76}  & --               & 77.9             & $+6.0\%$ \\
CWQ (Hit)           & 60.75\sstd{0.68}  & --               & 67.3             & $+10.8\%$ \\
CWQ (H@1)           & 57.37\sstd{0.23}  & --               & 64.5             & $+12.4\%$ \\
CWQ (F1)            & 55.23\sstd{0.45}  & --               & 62.1             & $+12.4\%$ \\
\midrule
\multicolumn{5}{@{}l}{\textit{(3) Molecule Property Prediction}} \\
BACE (ROC-AUC)        & 82.0\sstd{1.2}  & --               & 79.8             & $-2.7\%$ \\
BBBP (ROC-AUC)        & 70.6\sstd{2.3}  & --               & 71.6             & $+1.4\%$ \\
HIV  (ROC-AUC)        & 76.9\sstd{1.7}  & --               & 75.5             & $-1.8\%$ \\
\midrule
\multicolumn{5}{@{}l}{\textit{(4) ChEBI-20 Molecule Captioning}} \\
BLEU-2            & 0.534\sstd{0.009} & --               & 0.572\sstd{0.003} & $+7.1\%$ \\
BLEU-4            & 0.437\sstd{0.011} & --               & 0.482\sstd{0.003} & $+10.3\%$ \\
ROUGE-L           & 0.551\sstd{0.006} & --               & 0.583\sstd{0.002} & $+5.8\%$ \\
METEOR            & 0.567\sstd{0.008} & --               & 0.602\sstd{0.002} & $+6.2\%$ \\
\bottomrule
\end{tabular}
\end{table}

The task-specific hyperparameters are highlighted in their appropriate Appendix sections.

\section{Performance Attribution}
\label{app:performance-attribution}

In order to isolate where the performance gains come from, we conduct a smaller experiment where we compare the standard GTLM architecture to a variant without structural attention biases ($\forall u, v: b(u,v)=0$). Table \ref{tab:text-only-ablation} outlines the results of this experiment, which helps us attribute the performance to the two mechanisms GTLM innovates with:
\begin{itemize}
    \item[1)] fixing the \textbf{compression bottleneck} -- most impactful in the TAG benchmarks, as the difference between text-only GTLM and standard GTLM is not as large;
    \item[2)] structural \textbf{attention biases} -- most impactful in tasks where the question asks about structure, as demonstrated with WebQSP F1 score and GraphQA.
\end{itemize}

\begin{table}[ht]
\centering
\caption{Comparison between standard GTLM and its text-only variant (without structural attention biases).}
\label{tab:text-only-ablation}
\small
\setlength{\tabcolsep}{14pt}
\begin{tabular}{@{}l c@{\hspace{2.6em}}c c@{\hspace{2.6em}}c c c@{}}
\toprule
 & \multicolumn{2}{c}{Cora} & \multicolumn{2}{c}{PubMed} & WebQSP & GraphQA \\
\cmidrule(lr){2-3} \cmidrule(lr){4-5} \cmidrule(lr){6-6} \cmidrule(l){7-7}
Method & Acc & F1 & Acc & F1 & F1 & Task-mean Acc \\
\midrule
GTLM {\scriptsize(text-only)} & 86.78\sstd{0.57} & 90.17\sstd{0.35} & 93.98\sstd{0.12} & 94.26\sstd{0.12} & 48.5 & 33.2 \\
GTLM {\scriptsize(standard)}  & 88.99\sstd{1.12} & 91.31\sstd{0.90} & 94.57\sstd{0.11} & 94.76\sstd{0.10} & 73.5 & 85.7 \\
\bottomrule
\end{tabular}
\end{table}

\section{Simulated Message Passing---Experiment Details}
\label{app:simulated-message-passing}

To investigate GTLM's internal learned mechanisms (as discussed in Section~\ref{sec:simulated-message-passing}), we designed a controlled synthetic connectivity task. This section outlines the high-level methodology used to isolate this emergent behavior, while full implementation details and hyperparameter configurations are available in our project codebase.

\subsection{Dataset Generation}
We programmatically generated a dataset of random graphs to evaluate the model's ability to deduce whether two arbitrary nodes belong to the same clique. Each graph is generated with two or more isolated connected components. To prevent the model from relying on language priors, node attributes are restricted to randomly assigned single letters.

To query the model, we introduce the \textit{prompt node} containing the textual question (e.g., \texttt{"Are the nodes \{x\} and \{y\} connected? [Yes/No]"}). This prompt node is explicitly linked via directed edges to the two nodes in question. The dataset is strictly balanced to contain an equal number of positive examples (targets share a connected component) and negative examples (targets belong to disjoint components).

\subsection{Training and Evaluation Setup}
\textbf{Model Configuration.} We utilize \texttt{Llama-3.2-1B} as the base model, augmented with our full suite of structural attention biases (SPD, RRWP, and Magnetic Laplacian). To ensure parameter efficiency, the base model is frozen, and we \textbf{do not} apply any Low-Rank Adapters (LoRA). 

\textbf{Optimization Strategy.} We only train the randomly initialized bias-related parameters in our model. This setup allows us to only observe the effects of our newly introduced structure-aware attention biases on the model's behavior.

\textbf{Findings.} We find that the model learns to solve this task with close to $100\%$ accuracy. We further investigate the internal attention scores on one synthetic example and notice that multiple attention heads have learned to simulate message passing. For the visualization in Section~\ref{sec:simulated-message-passing} we used one specific attention head which best illustrated the point, while many other attention heads also exhibited a clear awareness of the graph structure---either through a similar ``message passing'' mechanism or its inverse where disconnected nodes have high attention scores, etc.

In summary, this experiment suggests that our GTLM architecture has the capacity for simulating the functionalities of dedicated GNNs and conventional GTs.

\section{Flex Attention with GTLM}
\label{app:flex-attention}

\subsection{Implementation}

The fused FlashAttention kernels are ineligible due to the additive per-head and per-layer graph bias. The eager implementation materializes the bias at token level, $(B, H, L, L)$, and this is what would bound the sequence lengths GTLM could train on---not the architecture. However, our backend keeps the bias at node level instead, storing only $[b(u,v)]_{u,v} \in \mathbb{R}^{N \times N}$ per head and layer and gathering the relevant entry inside the kernel through a sequence-to-node index $\pi: \{1,\dots,L\} \to \{1,\dots,N\}$, so the \texttt{score\_mod} is a single lookup:
\begin{verbatim}
    def score_mod(score, b, h, q, kv):
        return score + node_bias[b, h, node_id[q], node_id[kv]]
\end{verbatim}

Since the kernel never depends on how the \texttt{node\_bias} was produced, any bias type we define is compatible with these kernels, making the model's perception of graphs very easy to modify to experiment with different bias types. The backend requires $L$ to be padded to a multiple of the block size. Further, to overcome excessive kernel re-compiling, we pad the sequences and node features into a few dozen predefined $(N,L)$ buckets. Decoding uses a hybrid approach---due to the aforementioned requirement of FlexAttention to have discrete shape buckets---we route the prefill phase through our FlexAttention kernel, while each of the next autoregressive generation steps falls back to eager attention.

\subsection{Cost on the text-attributed-graph benchmarks}

Table~\ref{tab:flex-tag} shows training-step latency over $24$ randomly sampled real batches per benchmark. All arms receive identical tokens through the collator and the same padding method is used. We compare our own \textsc{flex} implementation to the naive eager backend for GTLM, as well as a standard LLM without any graph machinery. We find that \textsc{flex} is much faster and significantly more stable than the \textsc{eager} counterpart, while only $2\times$ slower than the LLM floor.

\begin{table}[h]
\centering
\small
\caption{Training-step latency on the TAG benchmarks (H100 80GB, bf16, $B=1$, LoRA, gradient checkpointing enabled). $L$ is the mean unpadded token count per graph.}
\begin{tabular}{lrrrrrr}
\toprule
 & & \multicolumn{3}{c}{step latency (ms)} & \multicolumn{2}{c}{ratio} \\
\cmidrule(lr){3-5}\cmidrule(lr){6-7}
dataset & $L$ & \textsc{eager} & \textsc{flex} & \textsc{flash} & eager/flex & flex/flash \\
\midrule
cora       &   663 &   836.9 & 226.8 & 118.6 & $3.69\times$ & $1.91\times$ \\
pubmed     &   950 & 1,255.3 & 215.9 & 112.1 & $5.81\times$ & $1.93\times$ \\
reddit     &   976 &   406.3 & 201.3 & 103.6 & $2.02\times$ & $1.94\times$ \\
ogbn-arxiv & 1,065 & 2,006.2 & 208.3 & 103.7 & $9.63\times$ & $2.01\times$ \\
\bottomrule
\end{tabular}
\label{tab:flex-tag}
\end{table}

The consistency arguably matters even more than some multiplier: \textsc{flex} lands at $201$--$227$\,ms and \textsc{flash} at $104$--$119$\,ms on every dataset, while only \textsc{eager} varies significantly, from $406$ to $2{,}006$\,ms. Peak memory falls with it ($6.33 \to 4.85$\,GB on cora, matching \textsc{flash}), and one epoch of ogbn-arxiv falls from $50.7$ to $5.3$ hours.

\subsection{Scaling}

\begin{figure}[h]
\centering
\includegraphics[width=\textwidth]{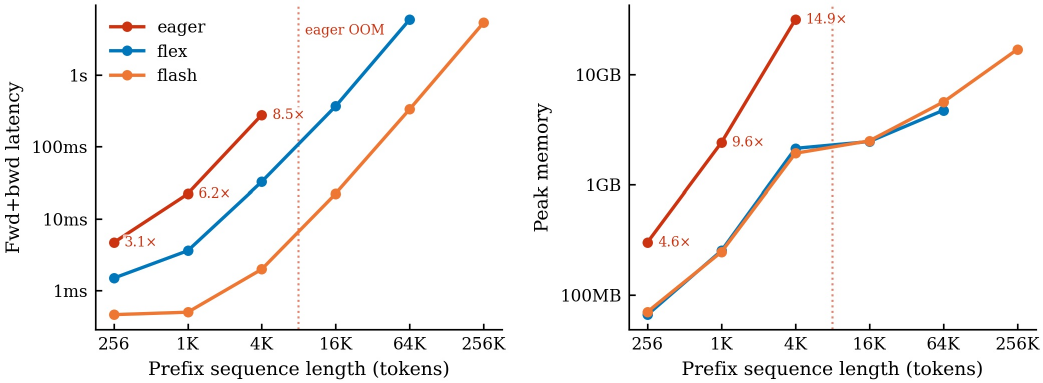}
\caption{Forward+backward latency (left) and peak memory (right) for one attention layer against packed sequence length, on synthetic graph batches with real topology and shortest-path distances. Annotations on the dense curve give its cost relative to ours, and the dotted line marks where it exhausts memory. Flash attention is shown as a bias-free lower bound; it cannot express the graph bias.}
\label{fig:flex-scaling}
\end{figure}

Figure~\ref{fig:flex-scaling} measures one layer of attention in isolation, where there is an $O(L^2)$ bias materialization. Relative to the dense path---the only baseline that computes the same function---our advantage scales with length: $3.1\times$, $6.2\times$, and $8.5\times$ less time, and $4.6\times$, $9.6\times$, and $14.9\times$ less memory, at $256$, $1$k, and $4$k tokens, respectively. Above $L \approx 4.5$k the dense path completely exhausts the 80\,GB device, while ours continues to $L \approx 70$k at less than $5$\,GB. Eager attention pays $O(L^2)$ memory per layer for the token-level bias; ours pays $O(N^2)$ on top of flash attention's $O(L)$, and $N \ll L$ since each node contributes many tokens.

Compilation is the only additional cost: one autotuned kernel per $(L, N)$ bin, $176$\,s on first encounter and $12$\,s from disk cache, over the four to ten bins that the TAG dataset produces. Amortized over tens of thousands of steps, it is negligible compared to the $2.0$-$9.6\times$ faster step per fraction of memory, at lengths that dense traversal cannot reach at all.

\section{Computational Resources and Hardware Setup}
\label{app:compute}

To ensure full reproducibility and provide transparency regarding the computational requirements of our architecture, we detail the hardware setup and empirical training times below.

Although some of our experiments used higher end graphics cards such as Nvidia's B200 or B300, all experiments can be conducted using a single NVIDIA A100 or H100 GPU (80GB VRAM). When fine-tuning the base \texttt{Llama-3.2-1B} model equipped with our structural LoRA adapters, training times on the individual GraphQA tasks ranged from 10 minutes (for standard graphs) to 40 minutes (for incidence graphs). The more demanding Text-Attributed Graph (TAG) benchmarks required between 6 and 8 hours of continuous training on an H100 GPU (with the exception of \textit{OGBN-Arxiv}, which required around 20 hours). Other experiments were carried out on those higher-end graphics cards, where training times ranged between 1 and 10 hours, depending on the task in question and the training configuration. Finally, training the 3B and 8B model variants introduced approximately $2\times$ and $3\times$ slowdowns, respectively.

\end{document}